\documentclass[10pt,twocolumn,letterpaper]{article}

\usepackage{cvpr}
\usepackage{times}
\usepackage{epsfig}
\usepackage{graphicx}
\usepackage{amsmath}
\usepackage{amssymb}
\usepackage{subfigure}
\usepackage{xcolor}
\usepackage{footnote}
\usepackage{makecell}

\usepackage{algorithm}
\usepackage{algpseudocode}

\pdfoutput=1

\graphicspath{{./figures/}{./figures/draw/}}

\def\x{{\mathbf x}}

\def\w{{\mathbf w}}
\def\h{{\mathbf h}}
\def\z{{\mathbf z}}
\def\cc{{\mathbf c}}
\def\bb{{\mathbf b}}
\def\I{{\mathbf 1}}
\def\q{{\mathbf q}}
\def\h{{\mathbf h}}
\def\f{{\mathbf f}}

\def\B{{\mathbf B}}

\def\X{{\mathbf X}}

\def\R{{\mathbb R}}
\def\H{{\mathbf H}}
\def\Z{{\mathbf Z}}
\def\W{{\mathbf W}}
\def\V{{\mathbf V}}
\def\Q{{\mathbf Q}}
\def\I{{\mathbf I}}
\def\1{{\mathbf 1}}

\def\V{{\mathbf V}}
\def\Vs{{\mathcal V}}

\def\0{{\mathbf 0}}

\newcommand\norm[1]{\left\lVert#1\right\rVert}
\newcommand{\overbar}[1]{\mkern 1mu\overline{\mkern-1mu#1\mkern-1mu}\mkern 1mu}

\usepackage[breaklinks=true,bookmarks=false]{hyperref}

\cvprfinalcopy 

\def\cvprPaperID{3012} 

\begin{document}

\title{Simultaneous Feature Aggregating and Hashing for Large-scale Image Search}

\author{Thanh-Toan Do$^{\dagger \star}$
\and
Dang-Khoa Le Tan$^{\star}$
\and Trung T. Pham$^{\dagger}$
\and
Ngai-Man Cheung$^{\star}$\\
\and $^{\dagger}$The University of Adelaide,  $^{\star}$Singapore University of Technology and Design \\
{\tt\small \{thanh-toan.do, trung.pham\}@adelaide.edu.au}, {\tt\small \{letandang\_khoa, ngaiman\_cheung\}@sutd.edu.sg}\\
}


    
\maketitle

\begin{abstract}

In most state-of-the-art hashing-based visual search systems, 
local image descriptors of an image are first aggregated as a single feature vector.
This feature vector is then subjected to a hashing function that produces a binary hash code. 
In previous work, the aggregating and the hashing processes are designed independently. 
In this paper, we propose a novel framework where feature aggregating and hashing are designed simultaneously and optimized jointly. 
Specifically, our joint optimization produces aggregated representations that can be better reconstructed by some binary codes.  This leads to 
more discriminative binary hash codes and improved retrieval accuracy.
In addition, we also 
propose a fast version of the recently-proposed Binary Autoencoder to be used in our proposed framework.
We perform 
extensive retrieval experiments on several benchmark datasets with both SIFT and convolutional features.
Our results suggest that the proposed framework achieves significant improvements over 
the state of the art.
\end{abstract}

\section{Introduction}
We are interested in the problem of large-scale image search in which finding a compact image representation is one of the crucial problems. State-of-the-art image search systems \cite{DBLP:conf/cvpr/JegouZ14,DBLP:conf/cvpr/ArandjelovicZ13,herve_cvpr2010,DBLP:conf/cvpr/ArandjelovicZ12,do_cvpr15} include three main steps in computing the image representation: local feature extraction, embedding, and aggregating. 
The local feature extraction step extracts a set of local features, e.g. SIFT \cite{SIFT_Lowe}, representing the image. The embedding step improves the discriminativeness of the local features by mapping these features into a high-dimensional space \cite{herve_cvpr2010,DBLP:conf/cvpr/JegouZ14,do_cvpr15}. The aggregating (pooling) step converts the set of mapped high dimensional vectors into a single vector representation which usually has the dimensionality of several thousands \cite{herve_cvpr2010,DBLP:conf/cvpr/JegouZ14,do_cvpr15}. In particular, the aggregating step is very important.  First, the aggregating step 
reduces the storage requirement which is one of main concerns in large-scale image search. Second, the aggregated representation vectors can be directly compared using standard metrics such as Euclidean distance.

Although the aggregated representation reduces the storage and allows simple distance-based comparison, it is not efficient enough for large-scale database which requires very compact representation and fast searching. An attractive approach for achieving these requirements is binary hashing. 
Specifically, binary hashing encodes the image representation into a compact binary hash code.
Existing binary hashing methods can be categorized as data-independent and data-dependent schemes \cite{DBLP:journals/corr/WangLKC15,DBLP:journals/corr/WangSSJ14,Grauman_review}. 
Data-dependent hashing methods use available training data for learning hash functions and they achieve better retrieval results than data-independent methods. The training can be unsupervised \cite{DBLP:conf/nips/WeissTF08,DBLP:conf/cvpr/GongL11,DBLP:conf/cvpr/HeWS13,CVPR12:SphericalHashing,do2016learning,do2016-sdp} or supervised \cite{DBLP:conf/icml/NorouziF11,Kulis_learningto,CVPR12:Hashing,CVPR2014Lin}. 
In particular, unsupervised hashing does not require any label information. Hence, it is suitable for large-scale image search in which the label information is usually unavailable. Therefore, our work focuses on the unsupervised hashing for large-scale image search.

In this work, we propose a novel framework where feature aggregating and hashing are designed simultaneously and optimized jointly. 
Traditionally, the aggregating/hashing processes are designed independently and separately \cite{DBLP:conf/cvpr/GongKRL13,DBLP:conf/cvpr/KimC15,DBLP:journals/pami/HeoLHCY15}:
First, some aggregation is applied on the local (embedded) features, resulting in a single aggregated representation for each image. Then, the set of aggregated representations is used for learning a hash function which encodes the aggregated representations into compact binary codes. 
For example, the recent 
Generalized Max Pooling \cite{DBLP:conf/cvpr/MurrayP14}
seeks a representation that can achieve some desirable aggregation property, i.e., equalizing the similarity between the representation and individual local features.
This aggregation process does not take into account any aspect of the subsequent hashing, and the resulted representations may not be suitable for hashing:   
in the context of unsupervised hashing, the aggregated representation may be difficult to be reconstructed by binary codes.
On the contrary, in our proposed simultaneous aggregating/hashing framework, we aim to compute aggregated representations that not only can achieve some desired aggregation property (equalized similarity) but also can be better reconstructed by some binary codes. As the aggregation is more reconstructible, the binary codes can retain more discriminative information, resulting in improved retrieval performance (in unsupervised hashing).

{\bf Our specific contributions are:} 
(i) To accelerate simultaneous learning of aggregating and hashing, we first propose a relaxed version of the state-of-the-art unsupervised hashing {Binary Autoencoder} \cite{BA_CVPR15} to be used in our framework. Instead of solving a NP-hard problem with the hard binary constraint on the outputs of the encoder, we propose to solve the problem with relaxation of the binary constraint, i.e., minimizing the binary quantization loss. In order to minimize this loss, we propose to solve the problem with alternating optimization. This proposed hashing method is not only faster in training but also competitive in retrieval accuracy when comparing to Binary Autoencoder \cite{BA_CVPR15}.
(ii) Our main contribution is a simultaneous feature aggregating/hashing learning approach which takes the local (embedded) features\footnote{In this work, the embedding is always applied when SIFT features are used.}
as inputs and learn the aggregation and hashing function simultaneously.
We propose alternating learning of the aggregated features and the hash function. 
(iii) The solid experiments on several image retrieval benchmark datasets show the proposed simultaneous learning significantly outperforms other recent unsupervised hashing methods. 

The remaining of this paper is organized as follows. Section \ref{sec:relatedwork} presents related works. Section \ref{sec:RBA} introduces the relaxed version of Binary Autoencoder \cite{BA_CVPR15}. Section \ref{sec:SAH} introduces the simultaneous feature aggregating and hashing. Section \ref{sec:evaSAH} presents experimental results. Section \ref{sec:concl} concludes the paper.

\section{Related work}
\label{sec:relatedwork}
\begin{table}[!t]
\footnotesize
\centering
\caption{Notations and their corresponding meanings.}
\label{tab:notation}
\begin{tabular}{|l|l|}
\hline
Notation	&Meaning \\ \hline
$\X$ &$\X = \{\x_i\}_{i=1}^{m} \in \R^{D\times m}$: set of $m$ training samples; \\
	 &each column of $\X$ corresponds to one sample\\ \hline	
$\Z$ &$\Z = \{\z_i\}_{i=1}^{m} \in \{-1,+1\}^{L\times m}$: binary code matrix \\ \hline
$L$  &Number of bits to encode a sample \\ \hline		
$\W_1, \cc_1$&$\W_1 \in \R^{L\times D}, \cc_1 \in \R^{L\times 1}$: weight and bias of encoder \\ \hline
$\W_2, \cc_2$&$\W_2 \in \R^{D\times L}, \cc_2 \in \R^{D\times 1}$: weight and bias of decoder \\ \hline
$\Vs$ &$\Vs=\{\V_i\}_{i=1}^m$; $\V_i\in \R^{D\times n_i}$ is set of local (embedded)\\ 
	 &representations of image $i$; \\
	 &$n_i$ is number of local descriptors of image $i$\\ \hline
$\Phi$ &$\Phi = \{\varphi_i\}_{i=1}^{m} \in \R^{D\times m}$: set of $m$ aggregated vectors; \\
	 &$\varphi_i$ corresponds to aggregated vector of image $i$\\ \hline	
$\1$ & column vector with all $1s$ elements\\ \hline
$\I$ & identity matrix \\ \hline
\end{tabular}
\end{table}
We summarize the notations in Table~\ref{tab:notation}. Two main components of the proposed simultaneous learning are aggregating and hashing. For aggregating, we rely on the state-of-the-art {Generalized Max Pooling} \cite{DBLP:conf/cvpr/MurrayP14}. For hashing, we propose a relaxed version of Binary Autoencoder \cite{BA_CVPR15}. This section presents a brief overview of Generalized Max Pooling \cite {DBLP:conf/cvpr/MurrayP14} and Binary Autoencoder \cite{BA_CVPR15}.  
\vspace{-0.3cm}
\paragraph{Generalized Max Pooling (GMP) \cite{DBLP:conf/cvpr/MurrayP14}}
Max-pooling~\cite{DBLP:conf/cvpr/YangYGH09,DBLP:conf/icml/BoureauPL10} is a common aggregation method which aggregates a set of local (embedded) vectors of the image to a single vector. 
However, classical max-pooling approach can only be applied to BoW or sparse coding features. 
Recently, in \cite{DBLP:conf/cvpr/JegouZ14} and \cite{DBLP:conf/cvpr/MurrayP14} the authors introduced a generalization of max-pooling (i.e., Generalized Max Pooling (GMP) \cite{DBLP:conf/cvpr/MurrayP14})\footnote{In \cite{DBLP:conf/cvpr/JegouZ14}, the authors named their method as \textit{democratic aggregation}. It actually shares similar idea to \textit{generalized max pooling} \cite{DBLP:conf/cvpr/MurrayP14}} which can be applied to general features such as VLAD \cite{jegou11d}, Temb \cite{DBLP:conf/cvpr/JegouZ14}, Fisher vector \cite{DBLP:conf/cvpr/PerronninD07}. The main idea of GMP is to equalize the similarity between each local embedded vector and the aggregated representation. In \cite{DBLP:conf/cvpr/JegouZ14,do_cvpr15}, the authors  showed that GMP achieves better retrieval accuracy than sum-pooling.
Given $\V \in \R^{D\times n}$, the set of $n$ embedded vectors of an image (each embedded vector has dimensionality $D$), GMP finds the aggregated representation $\varphi$ which equalizes the similarity (i.e. the dot-product)
 between each column of $\V$ and $\varphi$ by solving the following optimization
\begin{equation}
\min_{\varphi}\left(\norm{\V^T\varphi-\1}^2+\mu\norm{\varphi}^2 \right) \label{eq:gmp}
\end{equation}
(\ref{eq:gmp}) is a ridge regression problem which solution is 
\begin{equation}
\varphi = \left( \V\V^T + \mu\I\right)^{-1} \V\1 \label{eq:gmp-solution}
\end{equation}
\vspace{-0.9cm}
\paragraph{Binary Autoencoder (BA)\cite{BA_CVPR15}} 
In \cite{BA_CVPR15}, in order to compute the binary code, the authors minimize the following optimization
\vspace{-0.3cm}
\begin{equation}
\min_{\h,\f,\Z} \sum_{i=1}^m\left(\norm{\x_i - \f(\z_i)}^2+\mu\norm{\z_i-\h(\x_i)}^2 \right)
\label{eq:ba}
\end{equation}
\vspace{-0.2cm}
\begin{equation}
\textrm{s.t.\ } \z_i \in \{-1,1\}^{L}, i = 1,...,m \label{eq:ba-constraint}
\end{equation}
where $\h=sgn(\W_1\x + \cc_1)$ and $\f$ are encoder and decoder, respectively. By having $sgn$, the encoder will output binary codes. In the training of BA, the authors compute each variable $\f,\h,\Z$ at a time while holding the other fixed.
The authors show that the BA outperforms state-of-the-art unsupervised  hashing methods. 
However, the disadvantage of BA is time-consuming training which is mainly caused by the computing of $\h$ and $\Z$. As $\h$ involves $sgn$, it cannot be solved analytically. Hence, when computing $\h$, the authors cast the problem as the learning of $L$ separated linear SVM classifiers, i.e., for each $l=1,...,L$, they fit a linear SVM to $(\X, \Z_{l,.})$. When computing $\Z$, the authors solve for each sample $\x_i$ independently. 
Solving $\z_i$ in (\ref{eq:ba}) 
for each sample under binary constraint (\ref{eq:ba-constraint}) is NP-hard. To handle this, the authors first solve the problem with the relaxed constraint $\z_i \in [-1,1]$, resulting a continuous solution. 
They then apply the following procedure several times for getting $\z_i$: 
for each bit from $1$ to $L$, they evaluate the objective function when the bit equals $-1$ or $1$ with all remaining elements fixed and pick the best value for that bit. The asymptotic complexity for computing $\Z$ over all samples is $\mathcal{O}(mL^3)$.

In the following, we introduce our efficient Relaxed Binary Autoencoder algorithm (Section \ref{sec:RBA}) which will be used in our novel simultaneous feature aggregating and hashing framework (Section \ref{sec:SAH}). 
\section{Relaxed Binary Autoencoder (RBA)}
\label{sec:RBA}
\subsection{Formulation}
In order to achieve binary codes, we propose to solve the following constrained optimization

\vspace{-0.3cm}\footnotesize
\begin{eqnarray}
\min_{\{\W_i,\cc_i\}_{i=1}^{2}} J &=& \frac{1}{2} \norm{\X-\left(\W_2(\W_1\X+\cc_1\1^T)+\cc_2\1^T\right)}^2 \nonumber \\ 
{}&&+\frac{\beta}{2}\left(\norm{\W_1}^2+\norm{\W_2}^2\right) \label{eq:obj_ori}
\end{eqnarray}
\begin{equation}
\textrm{s.t. } \W_1\X+\cc_1\1^T \in \{-1,1\}^{L\times m} \label{eq:binary0}
\end{equation} 
\normalsize 
The constraint (\ref{eq:binary0}) makes sure the output of the encoder is binary. The first term of (\ref{eq:obj_ori}) makes sure the binary codes give a good reconstruction of the input, so it encourages (dis)similar inputs map to (dis)similar binary codes. The second term is a regularization that tends to decrease the magnitude of the weights, so it helps to prevent overfitting.

Solving (\ref{eq:obj_ori}) under (\ref{eq:binary0}) is difficult due to the binary constraint. In order to overcome this challenge, we propose to solve the relaxed version of the binary constraint, i.e., minimizing the binary quantization loss of the encoder. The proposed method is named as \textit{Relaxed Binary Autoencoder} (RBA). Specifically, we introduce a new auxiliary variable $\B$ and solve the following the optimization 

\vspace{-0.3cm}\footnotesize
\begin{eqnarray}
\min_{\{\W_i,\cc_i\}_{i=1}^{2},\B} J &=& \frac{1}{2} \norm{\X-\left(\W_2\B+\cc_2\1^T\right)}^2 \nonumber \\ 
{}&&\hspace{-8em}+\frac{\lambda}{2}\norm{\B-(\W_1\X+\cc_1\1^T)}^2+\frac{\beta}{2}\left(\norm{\W_1}^2+\norm{\W_2}^2\right) \label{eq:obj_2}
\end{eqnarray}
\begin{equation}
\textrm{s.t. } \B \in \{-1,1\}^{L\times m} \label{eq:binary2}
\end{equation} 
\normalsize 
The benefit of the auxiliary variable $\B$ is that we can decompose the difficult constrained optimization problem (\ref{eq:obj_ori}) into simpler sub-problems.  We use alternating optimization on these sub-problems as will be discussed in detail.


An important difference between the proposed RBA and the original BA is that our encoder does not involve $sgn$ function. The second term of (\ref{eq:obj_2}) forces the output of encoder close to binary values, i.e., it minimizes the binary quantization loss, while the first term still ensures good reconstruction loss. 
By setting the penalty parameter $\lambda$ sufficiently large, we penalize the binary constraint violation severely, thereby forcing the solution of (\ref{eq:obj_2}) closer to the feasible region of the original problem (\ref{eq:obj_ori}).

\subsection{Optimization}
In order to solve for $\W_1, \cc_1, \W_2, \cc_2$, $\B$ in (\ref{eq:obj_2}) under constraint (\ref{eq:binary2}), we solve each variable at a time while holding the other fixed.
\vspace{0.2cm}
\\
\textbf{$(\W,\cc)$-step:} When fixing $\cc_1,\cc_2$ and $\B$, we have the closed forms for $\W_1, \W_2$ as follows

\vspace{-0.2cm}\small
\begin{equation}
\W_1 = \lambda \left(\B-\cc_1\1^T\right)\X^T \left(\lambda\X\X^T+\beta\I\right)^{-1}
\label{eq:W1}
\end{equation}
\begin{equation}
\W_2 = \left(\X-\cc_2\1^T\right)\B^T \left(\B\B^T+\beta\I\right)^{-1}
\label{eq:W2}
\end{equation}
\normalsize 
When fixing $\W_1,\W_2$ and $\B$, we have the closed forms for $\cc_1, \cc_2$ as follows
\vspace{-0.2cm}
\begin{equation}
\cc_1 = \frac{1}{m}\left(\B-\W_1\X\right)\1 
\label{eq:c1}
\end{equation}
\begin{equation}
\cc_2 = \frac{1}{m}\left(\X-\W_2\B\right)\1
\label{eq:c2}
\end{equation}
\normalsize
Note that in (\ref{eq:W1}), the term $\X^T \left(\lambda\X\X^T+\beta\I\right)^{-1}$ is a constant matrix and it is computed only one time. 

\vspace{-0.3cm}
\paragraph{$\B$-step:} When fixing the weight and the bias, we can rewrite (\ref{eq:obj_2}) as
\begin{equation}
\norm{\widetilde{\X}-\W_2\B}^2 + \lambda \norm{\H-\B}^2
\end{equation}
\begin{equation}
\textrm{s.t. } \B \in \{-1,1\}^{L\times m} 
\end{equation} 
where $\widetilde{\X}=\X-\cc_2\1^T$ and $\H=\W_1\X+\cc_1\1^T$.

Inspired by the recent progress of discrete optimization \cite{Shen_2015_CVPR}, 
we use coordinate descent approach for solving $\B$, i.e., we solve one row of $\B$ each time while fixing all other rows. Specifically, let $\Q = \W_2^T\widetilde{\X} + \lambda \H$; for $k=1,...,L$, let $\w_k$ be $k^{th}$ column of $\W_2$; $\overbar{\W}_2$ be matrix $\W_2$ excluding $\w_k$; $\q_k$ be $k^{th}$ column of $\Q^T$; $\bb_k^T$ be $k^{th}$ row of $\B$; $\overbar{\B}$ be matrix $\B$ excluding $\bb_k^T$. We have the closed-form solution for $\bb_k^T$ as
\begin{equation}
\bb_k^T  = sgn \left(\q_k^T - \w_k^T\overbar{\W}_2\overbar{\B}\right)
\label{eq:bk}
\end{equation}
The proposed RBA is summarized in Algorithm~\ref{alg1}. In the Algorithm~\ref{alg1}, $\B^{(t)}$, $\W_1^{(t)}, \cc_1^{(t)}, \W_2^{(t)}, \cc_2^{(t)}$ are values at $t^{th}$ iteration. 
After learning ($\W_1, \cc_1, \W_2, \cc_2$), given a new vector $\x$, we pass $\x$ to the encoder, i.e., $h = \W_1\x+\cc_1$, and round the values of $h$ to $\{-1,1\}$, resulting binary codes. 
\begin{algorithm}[!t]
	\footnotesize
	\caption{Relaxed Binary Autoencoder (RBA)}
	\begin{algorithmic}[1] 
		\Require 
			\Statex $\X$: training data; $L$: code length; $T_1$: maximum iteration number; parameters $\lambda, \beta$
		\Ensure 
			\Statex 
			Parameters $\W_1, \cc_1, \W_2, \cc_2$
			\Statex 
			\State Initialize $\B^{(0)} \in \{−1,1\}^{L\times m}$ using ITQ~\cite{DBLP:conf/cvpr/GongL11}
			\State Initialize $\cc_1^{(0)}=\mathbf{0}$, $\cc_2^{(0)}=\mathbf{0}$ 
			\For{$t = 1 \to T_1$}
				\State Fix $\B^{(t-1)}, \cc_1^{(t-1)}, \cc_2^{(t-1)}$, solve $\W_1^{(t)}, \W_2^{(t)}$ by (\ref{eq:W1}) and (\ref{eq:W2}).
				\State Fix $\B^{(t-1)}, \W_1^{(t)}, \W_2^{(t)}$, solve $\cc_1^{(t)}, \cc_2^{(t)}$ by (\ref{eq:c1}) and (\ref{eq:c2}).
				\State Fix $\W_1^{(t)}, \W_2^{(t)}, \cc_1^{(t)}, \cc_2^{(t)}$, solve $\B^{(t)}$ by \textbf{B-step}.
			\EndFor
			\State Return 
			$\W_1^{(T_1)}, \W_2^{(T_1)}, \cc_1^{(T_1)}, \cc_2^{(T_1)}$
    \end{algorithmic}
    \label{alg1}
\end{algorithm}

\vspace{-0.3cm}
\paragraph{Comparison to Binary Autoencoder (BA) \cite{BA_CVPR15}:} There are two main advances of the proposed RBA (\ref{eq:obj_2}) over BA (\ref{eq:ba}). First, our encoder does not involve the $sgn$ function. Hence, during the iterative optimization, instead of using SVM for learning the encoder as in BA, we have an analytic solution ((\ref{eq:W1}) and (\ref{eq:c1})) for the encoder. Second, when solving for $\B$, 
instead of solving each sample at a time as in BA, we solve all samples at the same time by adapting the recent advance discrete optimization technique \cite{Shen_2015_CVPR}.
The asymptotic complexity for computing one row of $\B$, i.e. (\ref{eq:bk}), is $\mathcal{O}(mL)$. Hence the asymptotic complexity for computing $\B$ is only $\mathcal{O}(mL^2)$ which is less than $\mathcal{O}(mL^3)$ of BA. These two advances makes the training of RBA is faster than BA. 
\subsection{Evaluation of Relaxed Binary Autoencoder (RBA)}
This section evaluates the proposed RBA and compares it to the following state-of-the-art unsupervised hashing methods: 
 Iterative Quantization (ITQ)~\cite{DBLP:conf/cvpr/GongL11}, Binary Autoencoder (BA)~\cite{BA_CVPR15}, Spherical Hashing (SPH)~\cite{CVPR12:SphericalHashing}, K-means Hashing (KMH)~\cite{DBLP:conf/cvpr/HeWS13}. For all compared methods, we use the implementations and the suggested parameters provided by the authors. The values of $\lambda$, $\beta$ and the number of iteration $T_1$ in the Algorithm \ref{alg1} are empirically set by cross validation as $10^{-2}, 1$ and $10$, respectively. 
 The BA \cite{BA_CVPR15} and the proposed RBA required an initialization for the binary code. 
 To make a fair comparison, 
 we follow \cite{BA_CVPR15}, i.e., using ITQ \cite{DBLP:conf/cvpr/GongL11} for the initialization.
\vspace{-0.1cm}
\subsubsection{Dataset and evaluation protocol}
\label{subsub_eva_rba}
\vspace{-0.2cm}
\paragraph{Dataset} We conduct experiments on  CIFAR10 \cite{Krizhevsky09}, MNIST \cite{mnistlecun} and SIFT1M \cite{herve_pami2011} datasets which are widely used in evaluating hashing methods \cite{DBLP:conf/cvpr/GongL11,BA_CVPR15}.

{CIFAR10 dataset~\cite{Krizhevsky09}} consists of 60,000 images of 10 classes. 
The dataset is split into training and test sets, with $50,000$ and $10,000$ images,  respectively. Each image is represented by 320 dimensional GIST feature~\cite{gist}.

{MNIST dataset~\cite{mnistlecun}} consists of 70,000 handwritten digit
images of 10 classes. 
The dataset is split into training and test sets, with $60,000$ and $10,000$ images,  respectively. Each image is represented by a 784 dimensional gray-scale feature vector.

{SIFT1M dataset \cite{herve_pami2011}} contains 128 dimensional SIFT vectors \cite{SIFT_Lowe}. There are 1M vectors used as database for retrieval, 100K vectors for training, 
and 10K vectors for query. 
\vspace{-0.3cm}
\paragraph{Evaluation protocol}
In order to create ground truth for queries, we follow \cite{DBLP:conf/cvpr/GongL11,BA_CVPR15} in which the Euclidean nearest neighbors are used. The number of ground truths is set as in~\cite{BA_CVPR15}. For each query in CIFAR10 and MNIST datasets, its $50$ Euclidean nearest neighbors are used as ground truths; for each query in the large scale dataset SIFT1M, its 10,000  Euclidean nearest neighbors are used as ground truths. Follow the state of the art \cite{DBLP:conf/cvpr/GongL11,BA_CVPR15}, the performance of methods is measured by mAP. Note that as computing mAP is slow on the large scale dataset SIFT1M, we consider top 10,000 returned neighbors when computing mAP.
\vspace{-0.2cm}
\subsubsection{Experimental results}
\begin{figure}[!t]
\centering
\subfigure[CIFAR10]{
       \includegraphics[scale=0.27]{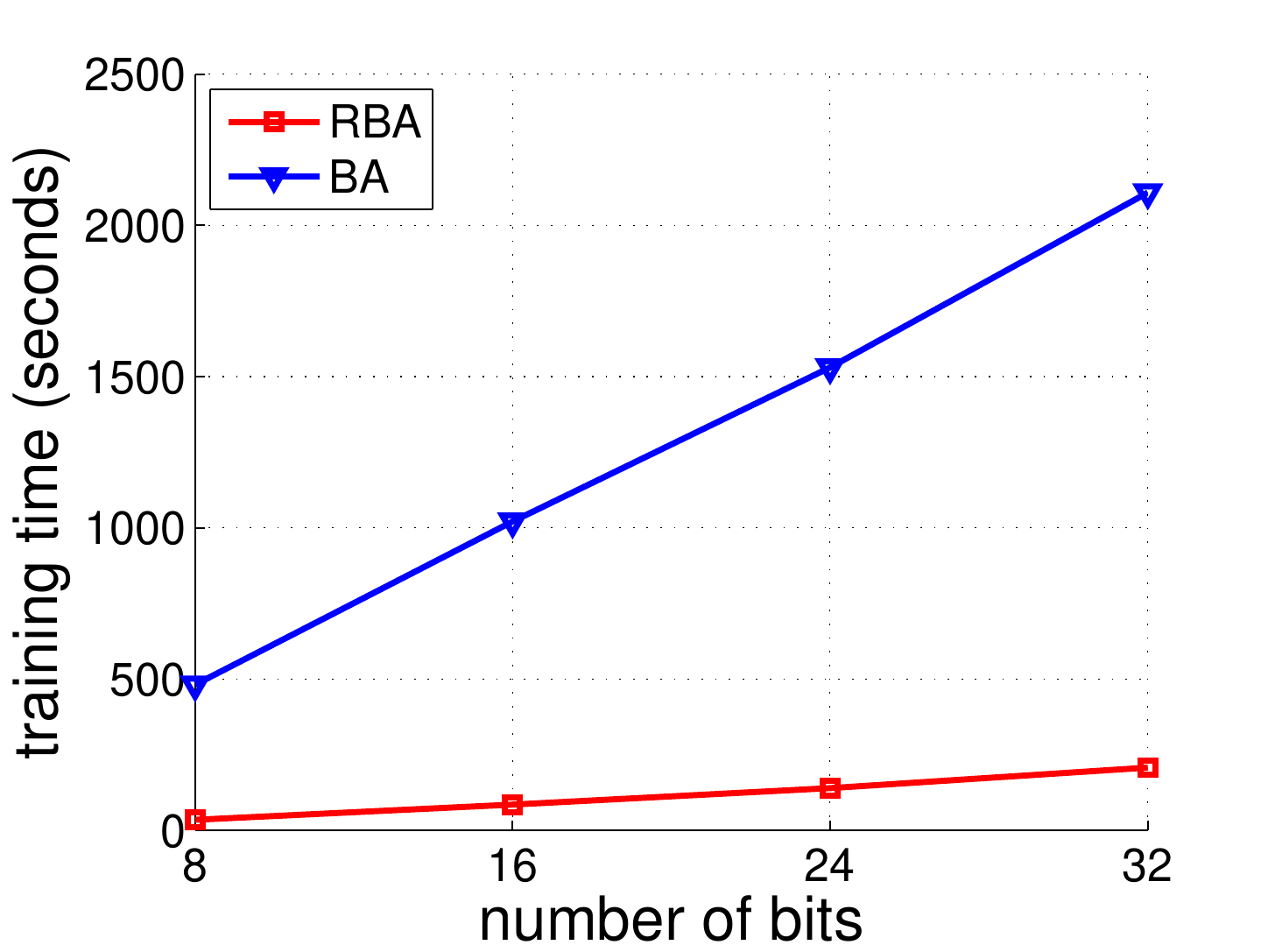}
       \label{fig:cifar_time}
}
\hspace{-0.2cm}
\subfigure[SIFT1M]{
       \includegraphics[scale=0.27]{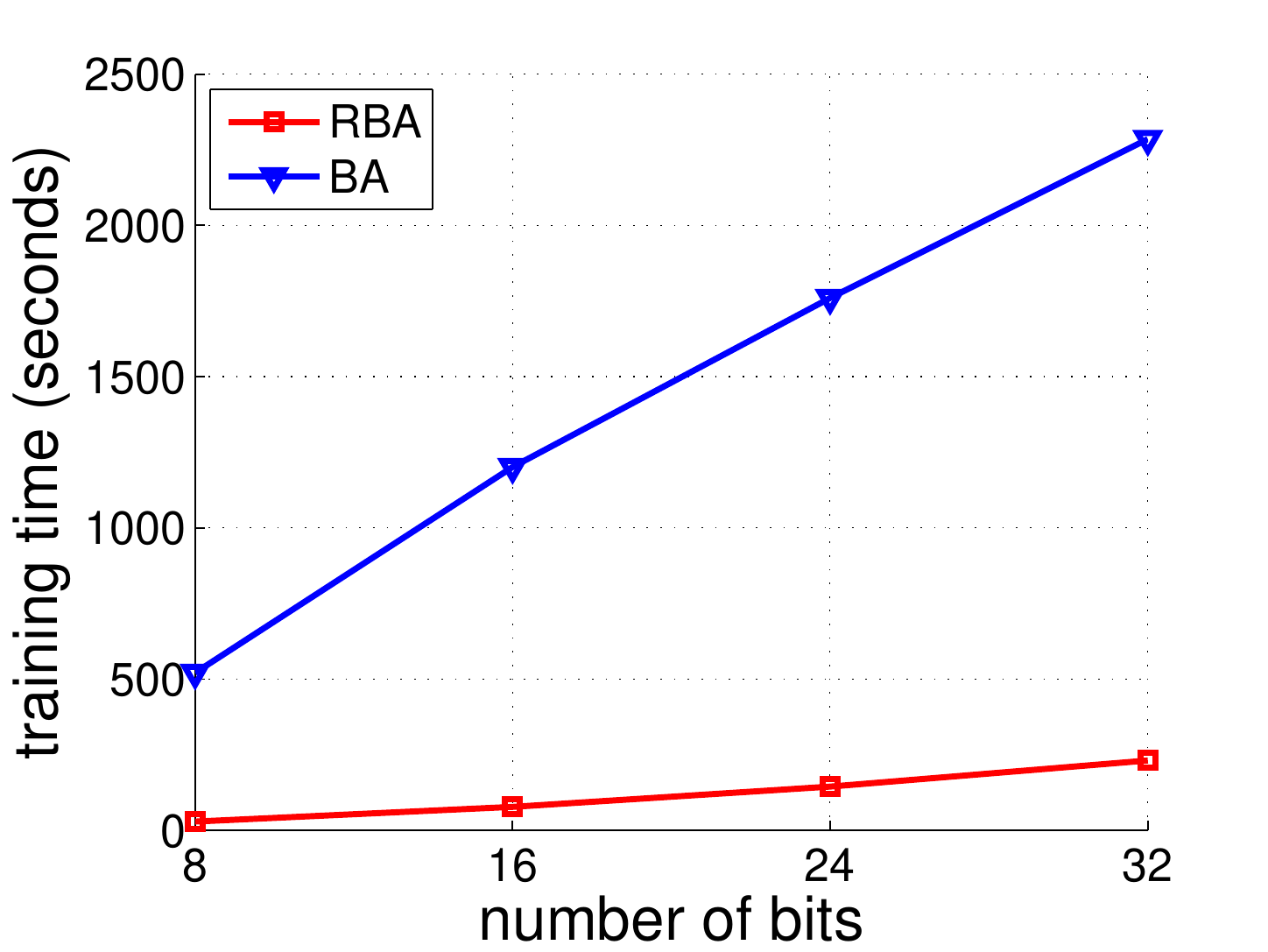} 
       \label{fig:sift_time}
}
\caption[]{Training time of BA and RBA on CIFAR10 and SIFT1M}
\label{fig:time-rba-ba}
\end{figure}

\vspace{-0.2cm}
\paragraph{Training time of RBA and BA}
In this experiment, we empirically compare the training time of RBA and BA. The experiments are carried out on a processor core (Xeon E5-2600/2.60GHz). It is worth noting that the implementation of RBA is in Matlab, while BA 
optimizes the implementation by using mex-files 
at the encoder learning step. The comparative training time on CIFAR10 and SIFT1M datasets is showed in Figure \ref{fig:time-rba-ba}. The results show that RBA is more than ten times faster training than BA for all code lengths on both datasets. The training time of BA is almost linear to the number of bits. This can be explained as follows: the most training time of BA is to solve the encoder and $\Z$. For both problems, they solve each bit separately (Section \ref{sec:relatedwork}), i.e., for encoder, they learn $L$ SVMs; for $\Z$, they check the optimum value of each bit sequentially. 
%

\vspace{-0.3cm}
\paragraph{Retrieval results}
\begin{figure*}[!t]
\centering
\vspace{-0.3cm}
\subfigure[CIFAR10]{
       \includegraphics[scale=0.33]{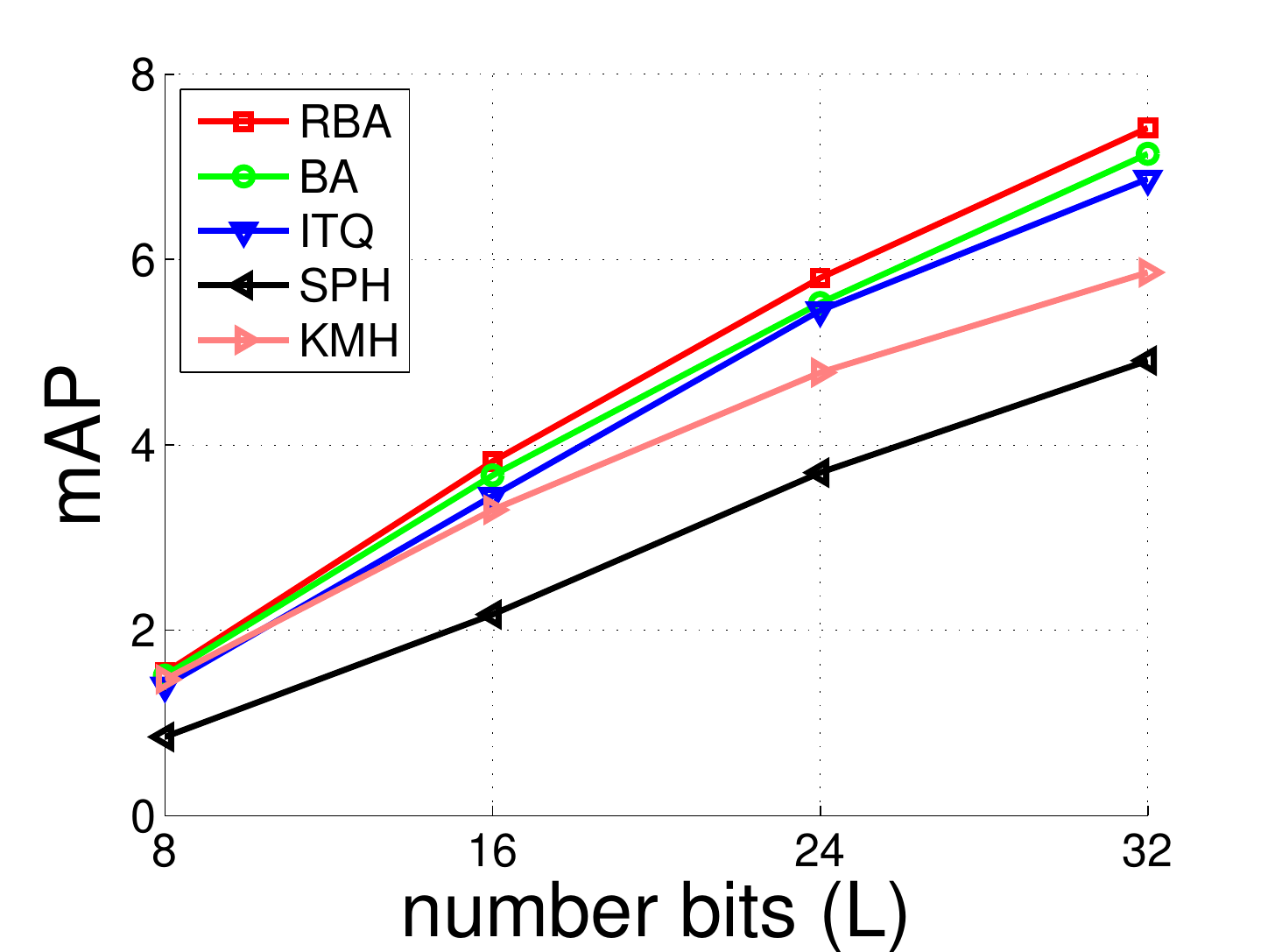}
       \label{fig:cifar_mAP}
}
\subfigure[MNIST]{
       \includegraphics[scale=0.33]{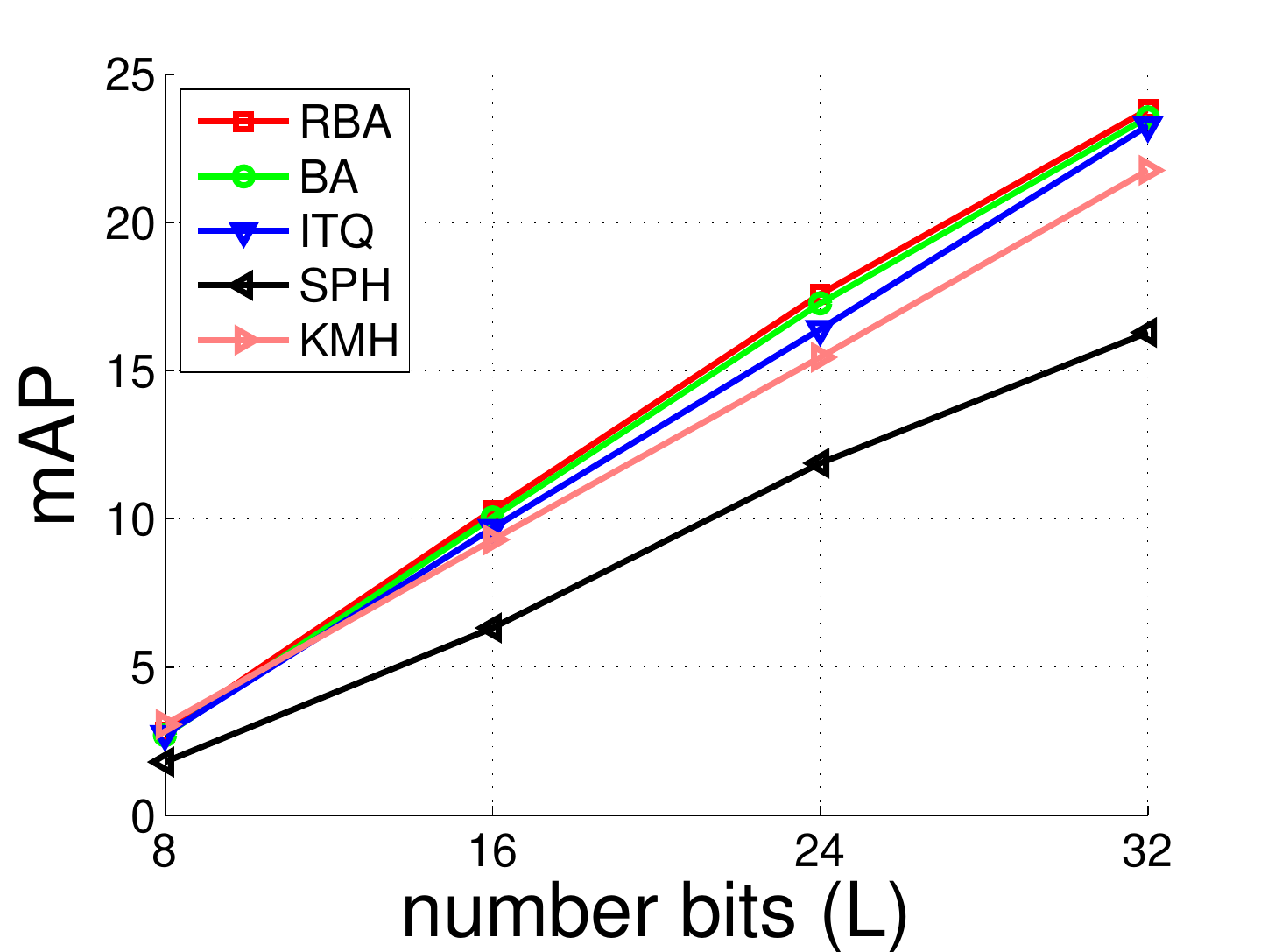} 
       \label{fig:mnist_mAP}
}
\subfigure[SIFT1M]{ 
       \includegraphics[scale=0.33]{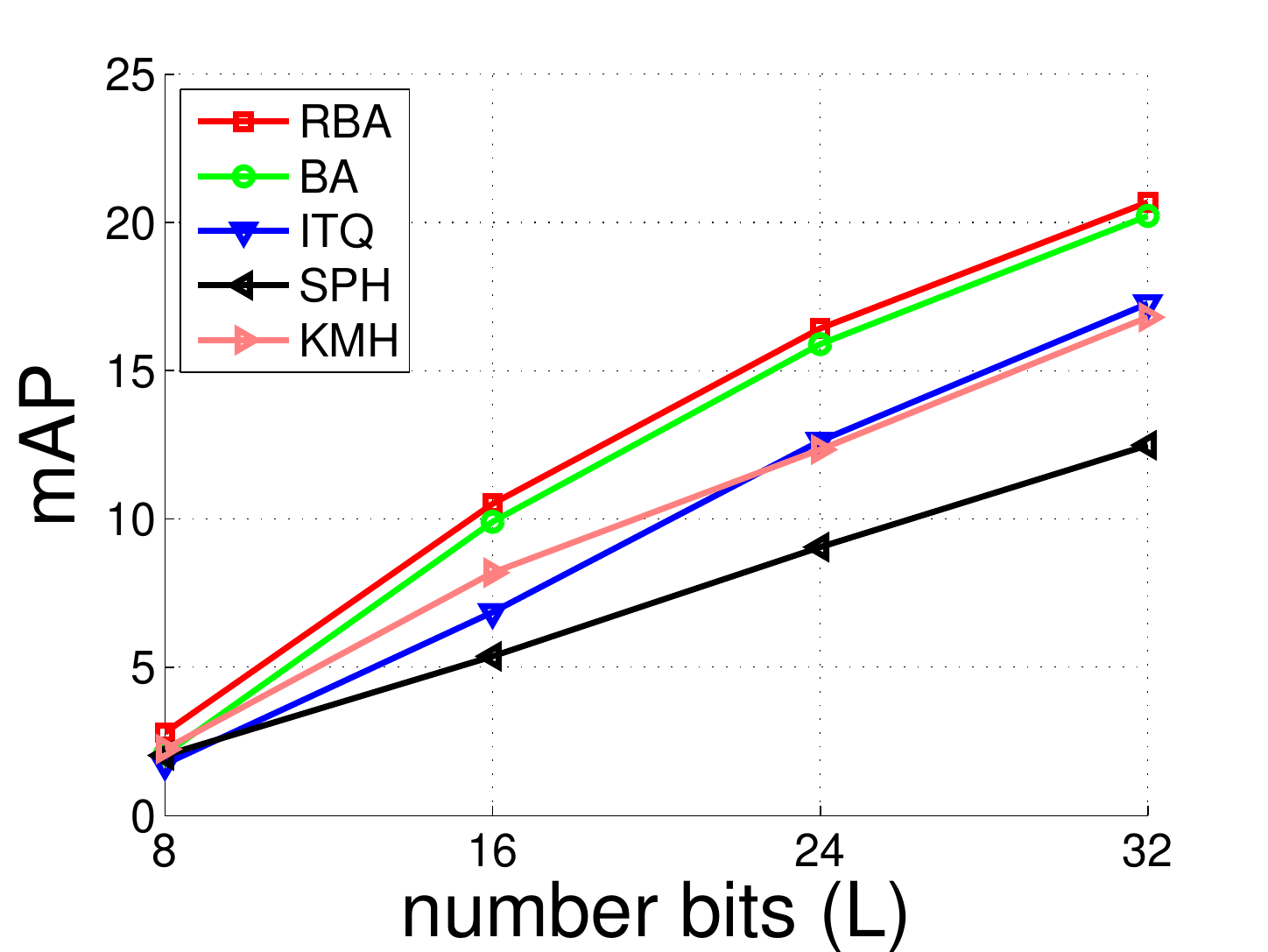} 
       \label{fig:sift1m_mAP}
}
\caption[]{mAP comparison between RBA and state-of-the-art unsupervised hashing methods on CIFAR10, MNIST, and SIFT1M}
\label{fig:rba-soa}
\end{figure*}

Figure \ref{fig:rba-soa} shows the comparative mAP between methods. We find the following observations are consistent for all three datasets. At all code lengths, the proposed RBA outperforms or is competitive with the state-of-the-art BA. This result confirms the advance of our approach for computing encoder (i.e., closed-form) and \textbf{$\B$-step} 
(i.e. using coordinate descent with closed-form for each row).
The results in Figure \ref{fig:rba-soa} also confirm the superior performance of BA and RBA over other methods. The improvements are more clear on the large scale SIFT1M dataset.

\section{Simultaneous Feature Aggregating and Hashing (SAH)}
\label{sec:SAH}
\subsection{Formulation}
Our goal is to simultaneously learn the aggregated vector representing an image and the hashing function, given the set of local image representations. 
For simultaneous learning, the learned aggregated vectors and the hash parameters should ensure desired properties of both aggregating and hashing.  Specifically, {\em aggregating property}: (i) for each image $i$, the dot-product similarity between the aggregated vector $\varphi_i$ and each local vector of $\V_i$ should be a constant; {\em hashing properties}: (ii) the outputs of the encoder are binary and (iii) the binary codes should preserve the similarity between image representations. In order to achieve these properties, we formulate the simultaneous learning as the following optimization

\vspace{-0.3cm}\footnotesize
\begin{eqnarray}
{}&&\hspace{-3em}\min_{\W_1,\cc_1,\W_2,\cc_2,\Phi} \frac{1}{2} \norm{\Phi-\left(\W_2(\W_1\Phi+\cc_1\1^T)+\cc_2\1^T\right)}^2 \nonumber \\ 
{}&&\hspace{-4em}+\frac{\beta}{2}\left(\norm{\W_1}^2+\norm{\W_2}^2\right)+\frac{\gamma}{2}\sum_{i=1}^{m}\left(\norm{\V_i^T\varphi_i-\1}^2+\mu\norm{\varphi_i}^2 \right) \label{eq:obj_ori_join}
\end{eqnarray}
\begin{equation}
\textrm{s.t. } \W_1\Phi+\cc_1\1^T \in \{-1,1\}^{L\times m} \label{eq:binary_join}
\end{equation} 
\normalsize 
The first term of (\ref{eq:obj_ori_join}) ensures a good reconstruction of $\Phi$, hence it encourages the similarity preserving (the property iii). The binary constraint (\ref{eq:binary_join}) ensures the binary outputs of encoder (the property ii). Finally, the third term ensures the learned aggregated representation equals the similarities between $\varphi_i$ and different columns of $\V_i$ by forcing their inner product to be $1$ (the property i). 

\subsection{Optimization}
In order to solve (\ref{eq:obj_ori_join}) under constraint (\ref{eq:binary_join}), we propose to iteratively optimize it by alternatingly optimizing w.r.t. hashing parameters ($\W,\cc$) and  aggregated representation $\Phi$ while holding the other fixed. 
\vspace{-0.3cm}
\paragraph{$\Phi$-step:} When fixing $\W_1,\cc_1,\W_2,\cc_2$ and solving for $\Phi$, we can solve over each $\varphi_i$ independently. Specifically, for each sample $i=1,...,m$, we solve the following relaxed problem by skipping the binary constraint
\begin{eqnarray}
\min_{\varphi_i} \frac{1}{2} \norm{\varphi_i-\left(\W_2(\W_1\varphi_i+\cc_1)+\cc_2\right)}^2 \nonumber \\ 
{}&&\hspace{-17em}+\frac{\gamma}{2}\left(\norm{\V_i^T\varphi_i-\1}^2+\mu\norm{\varphi_i}^2 \right) \label{eq:obj_pool}
\end{eqnarray}
By solving (\ref{eq:obj_pool}), we find $\varphi_i$ which satisfies the properties (i) and (ii), i.e., $\varphi_i$ not only ensures the aggregating property but also minimize the reconstruction error w.r.t. the fixed hashing parameters. (\ref{eq:obj_pool}) is actually a $l_2$ regularized least squares problem, hence we achieve the analytic solution as

\vspace{-0.3cm}\footnotesize
\begin{eqnarray}
\varphi_i &=& \left((\I-\W_2\W_1)^T(\I-\W_2\W_1)+\gamma\V_i\V_i^T+\gamma\mu\I\right)^{-1} \nonumber \\ 
{}&&\hspace{0em}\times \left(\gamma\V_i\1+(\I-\W_2\W_1)^T(\W_2\cc_1+\cc_2)\right) \label{eq:sol_pool}
\end{eqnarray}
\normalsize
The asymptotic complexity for computing (\ref{eq:sol_pool}) is 
$\mathcal{O} (\max(D^3,D^2n_i))$ 
which is similar to the asymptotic complexity for computing (\ref{eq:gmp-solution}).
\vspace{-0.3cm}
\paragraph{$(\W,\cc)$-step:} When fixing $\Phi$ and solving for $(\W_1,\cc_1,\W_2,\cc_2)$, (\ref{eq:obj_ori_join}) under the constraint (\ref{eq:binary_join}) is equivalent to the following optimization

\vspace{-0.2cm}\footnotesize
\begin{eqnarray}
\min_{\{\W_i,\cc_i\}_{i=1}^{2}}\frac{1}{2} \norm{\Phi-\left(\W_2(\W_1\X+\cc_1\1^T)+\cc_2\1^T\right)}^2 \nonumber \\ 
{}&&\hspace{-20em}+\frac{\beta}{2}\left(\norm{\W_1}^2+\norm{\W_2}^2\right) \label{eq:obj_hash}
\end{eqnarray}
\begin{equation}
\textrm{s.t. } \W_1\Phi+\cc_1\1^T \in \{-1,1\}^{L\times m} \label{eq:binary-hash}
\end{equation} 

\normalsize 
By solving (\ref{eq:obj_hash}) under the constraint (\ref{eq:binary-hash}), we find hash parameters which satisfy the properties (ii) and (iii), i.e., they not only ensure the binary outputs of the encoder but also minimize the reconstruction error w.r.t. the fixed aggregated representation $\Phi$.
(\ref{eq:obj_hash}) and (\ref{eq:binary-hash}) have same forms as (\ref{eq:obj_ori}) and (\ref{eq:binary0}), so we solve this optimization with the proposed Relaxed Binary Autoencoder (Section \ref{sec:RBA}). We use the Algorithm \ref{alg1} for solving $(\W_1,\cc_1,\W_2,\cc_2)$ in which $\Phi$ is used as the training data.

The proposed simultaneous feature aggregating and hashing is presented in the Algorithm \ref{alg2}. In the Algorithm~\ref{alg2}, $\Phi^{(t)}$, $\W_1^{(t)}, \cc_1^{(t)}, \W_2^{(t)}, \cc_2^{(t)}$ are values at $t^{th}$ iteration. After learning $\W_1, \cc_1, \W_2, \cc_2$, given set of local features of a new image, we first compute its aggregated representation $\varphi$ using (\ref{eq:sol_pool}). We then pass $\varphi$ to the encoder 
to compute the binary codes. 


\begin{algorithm}[!t]
	\footnotesize
	\caption{Simultaneous feature Aggregating and Hashing (SAH)}
	\begin{algorithmic}[1] 
		\Require 
			\Statex $\Vs=\{\V_i\}_{i=1}^{m}$: training data; $L$: code length; $T, T_1$: maximum iteration numbers for SAH and RBA (Algorithm \ref{alg1}), respectively; parameters $\lambda, \beta, \gamma, \mu$. 
		\Ensure 
			\Statex 
			Parameters $\W_1, \cc_1, \W_2, \cc_2$
			\Statex 
			\State Initialize $\Phi^{(0)}=\{\varphi_i\}_{i=1}^{m}$ with Generalized Max Pooling (\ref{eq:gmp-solution})
			\For{$t = 1 \to T$}
				\State Fix $\Phi^{(t-1)}$, solve $(\W_1^{(t)}, \cc_1^{(t)}, \W_2^{(t)}, \cc_2^{(t)})$ using Algorithm \ref{alg1} 
				\State Fix $(\W_1^{(t)}, \cc_1^{(t)}, \W_2^{(t)}, \cc_2^{(t)})$, solve $\Phi^{(t)}$ using \textbf{$\Phi$-step}.
			\EndFor
			\State Return 
			$\W_1^{(T)}, \W_2^{(T)}, \cc_1^{(T)}, \cc_2^{(T)}$
    \end{algorithmic}
    \label{alg2}
\end{algorithm}

\section{Evaluation of Simultaneous Feature Aggregating and Hashing (SAH)}
\label{sec:evaSAH}
This section evaluates and compares the proposed SAH to the following state-of-the-art unsupervised hashing methods: 
 Iterative Quantization (ITQ)~\cite{DBLP:conf/cvpr/GongL11}, Binary Autoencoder (BA)~\cite{BA_CVPR15} and the proposed RBA, Spherical Hashing (SPH)~\cite{CVPR12:SphericalHashing}, K-means Hashing (KMH)~\cite{DBLP:conf/cvpr/HeWS13}. For all compared methods, we use the implementations and the suggested parameters provided by the authors. The values of $\lambda$, $\beta$, $\gamma$, and $\mu$ 
  are set by cross validation as $10^{-2}$, $10^{-1}$, $10$, and $10^2$, respectively. 
\vspace{-0.1cm}
\subsection{Dataset}
We conduct experiments on Holidays \cite{herve_ijcv2010} and Oxford5k \cite{Philbin07-cvpr-2007} datasets which are widely used in evaluating image retrieval systems \cite{DBLP:conf/cvpr/JegouZ14,DBLP:conf/cvpr/ArandjelovicZ13,herve_cvpr2010,do_cvpr15}.

\textbf{Holidays} The Holidays dataset  \cite{herve_ijcv2010} consists of 1,491 images of different locations and objects, 500 of them being used as queries. Follow \cite{DBLP:conf/cvpr/JegouZ14,do_cvpr15}, when evaluating, we remove the query from the ranked list. For the training dataset, we follow \cite{DBLP:conf/cvpr/JegouZ14,do_cvpr15}, i.e., using 10k images from the independent dataset Flickr60k provided with Holidays.

\textbf{Holidays+Flickr100k} In order to evaluate the proposed method on large scale, we merge Holidays dataset with 100k images downloaded from Flickr \cite{herve_eccv2008}, forming the Holidays+Flickr100k dataset. 
 This dataset uses the same training dataset with Holidays. 

\textbf{Oxford5k} 
The Oxford5k dataset \cite{Philbin07-cvpr-2007} consists of 5,063 images of buildings and 55 query images corresponding to 11 distinct buildings in Oxford. We follow standard protocol \cite{DBLP:conf/cvpr/JegouZ14,DBLP:conf/cvpr/ArandjelovicZ13}: the bounding boxes of the region of interest are cropped and then used as the queries. As standardly done in the literature, for the learning, we use the Paris6k dataset \cite{Philbin08_cvpr}.

The ground truth of queries have been provided with the datasets \cite{herve_ijcv2010,Philbin07-cvpr-2007}. Follow the state of the art \cite{DBLP:conf/cvpr/GongL11,BA_CVPR15}, we evaluate the performance of methods with mAP. 
\subsection{Experiments with SIFT features}
Follow state-of-the-art image retrieval systems \cite{DBLP:conf/cvpr/JegouZ14,herve_cvpr2010,do_cvpr15}, to describe images, we extract SIFT local descriptors \cite{SIFT_Lowe} on Hessian-affine regions \cite{mikolajczyk_scale_2004}. RootSIFT variant \cite{DBLP:conf/cvpr/ArandjelovicZ12} is used in all our experiments. 
Furthermore, instead of directly using SIFT local features,
as a common practice, 
we enhance their discriminative power by embedding them into high dimensional space (i.e., 1024 dimensions) with the state-of-the-art triangulation embedding \cite{DBLP:conf/cvpr/JegouZ14}.  As results, the set of triangulation embedded vectors $\Vs=\{\V_i\}_{i=1}^m$ is used as the input for the proposed SAH. 
In order to make a fair comparison to other methods, we aggregate the triangulation embedded vectors  with GMP \cite{DBLP:conf/cvpr/MurrayP14} and use the resulted vectors 
as input for compared hashing methods. 


\vspace{-0.3cm}
\paragraph{Reconstruction comparison}
In this experiment, we evaluate the reconstruction capacity of binary codes produced by different methods: ITQ \cite{DBLP:conf/cvpr/GongL11}, BA \cite{BA_CVPR15}, RBA,  and SAH. We compute the average reconstruction error on the Oxford5k dataset. 





For ITQ, BA, and RBA, given the binary codes $\Z$ of the testing data (Oxford5k), the reconstructed testing data is computed by $\X_{res} = \W_2 \Z + \cc_2 \1^T$, where ($\W_2, \cc_2$) is  decoder. Note that the decoder is available in the design of BA/RBA and is learned in learning process. For ITQ, there is no decoder in its design, hence we follow \cite{BA_CVPR15}, i.e., we compute the optimal linear decoder ($\W_2, \cc_2$) using the binary codes of the training data (Paris6k). 

For SAH, given the binary codes $\Z$, we use the learned encoder and decoder to compute the aggregated representations $\Phi$ by using (\ref{eq:sol_pool}). The reconstruction of $\Phi$ is computed by using the decoder as $\Phi_{res} = \W_2 \Z + \cc_2 \1^T$.
\begin{figure}[!t]
\centering
\includegraphics[height=4cm, width=6cm]{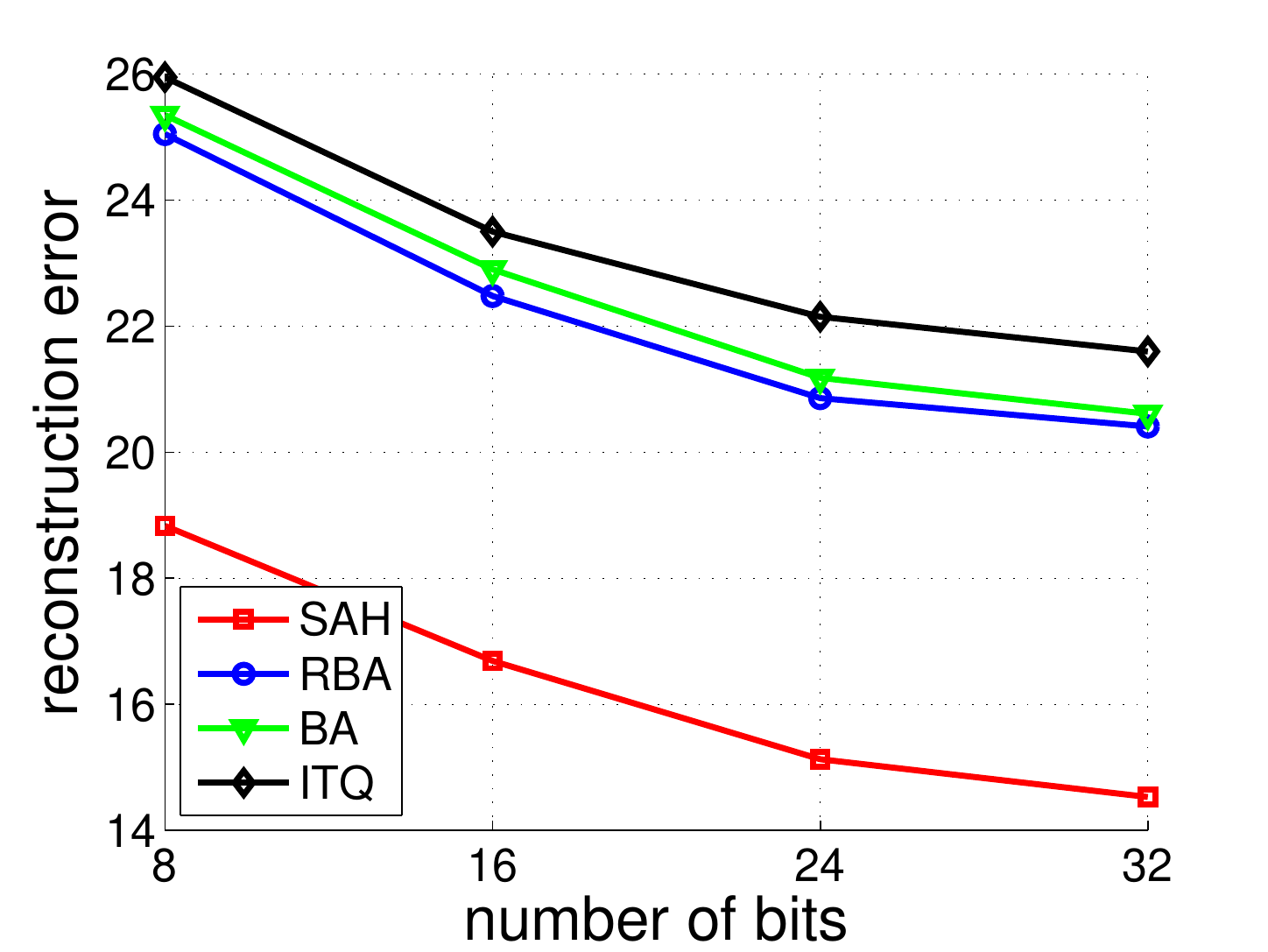}
\caption[]{Reconstruction error comparison of different methods on Oxford5k dataset}
\label{fig:res_oxford5k}
\end{figure}

Figure \ref{fig:res_oxford5k} shows that BA and RBA are comparable while SAH dominates all other methods in term of reconstruction error. This confirms the benefit of the jointly learning of aggregating and hashing in the proposed SAH. 
\vspace{-0.3cm}
\paragraph{Retrieval results}
\begin{figure*}[!t]
\centering
\vspace{-0.2cm}
\subfigure[Holidays]{
       \includegraphics[scale=0.33]{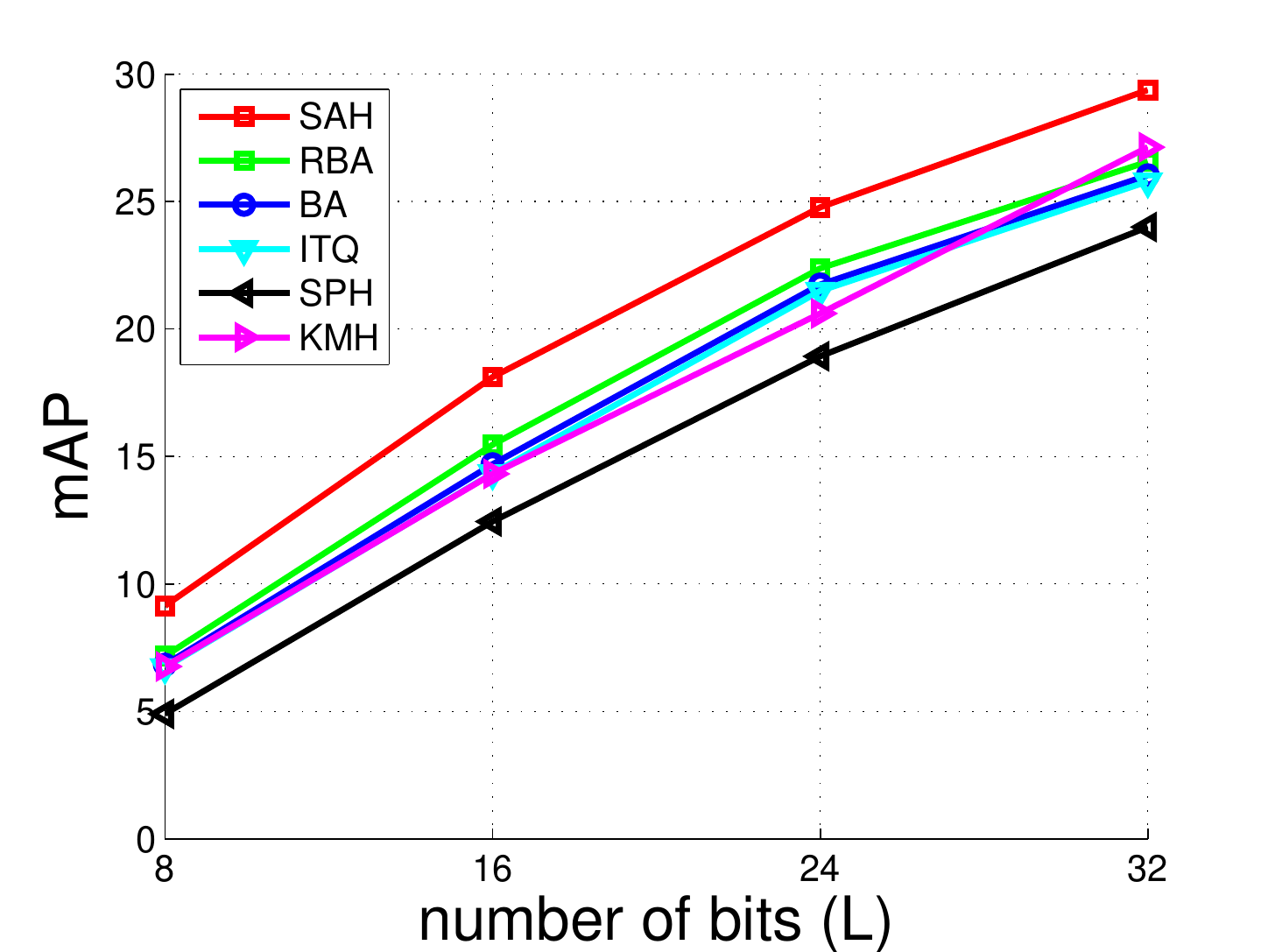}
       \label{fig:holidays_mAP_SIFT}
}
\subfigure[Oxford5k]{
       \includegraphics[scale=0.33]{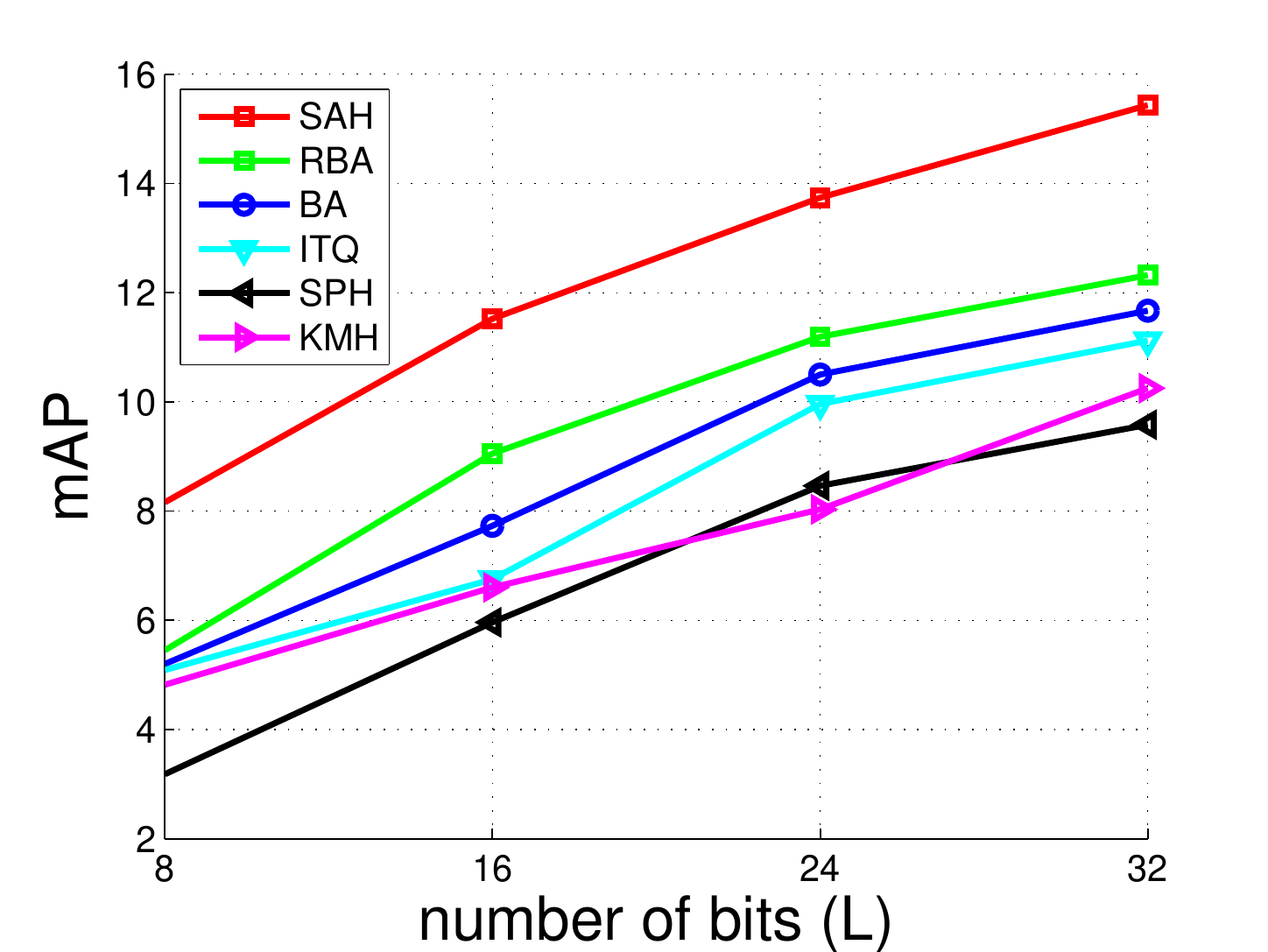} 
       \label{fig:oxford5k_mAP_SIFT}
}
\subfigure[Holidays+Flickr100k]{ 
       \includegraphics[scale=0.33]{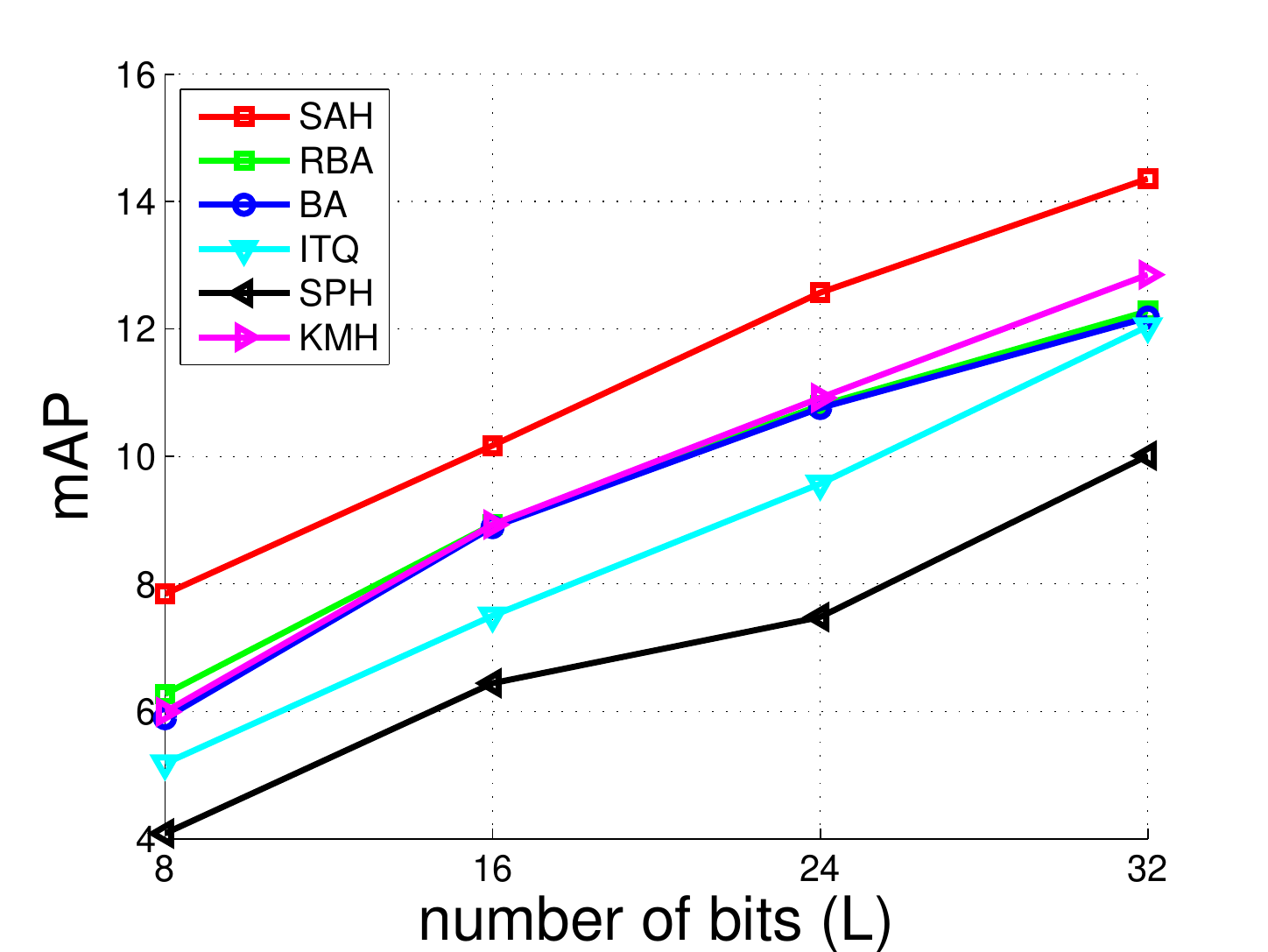} 
       \label{fig:holidays100k_mAP_SIFT}
}
\caption[]{mAP comparison between SAH and state-of-the-art unsupervised hashing methods when using SIFT features on Holidays, Oxford5k, and Holidays+Flickr100k}
\label{fig:lph-soa}
\end{figure*}

Figure \ref{fig:lph-soa} shows the comparative mAP between compared methods. We find the following observations are consistent on three datasets. The proposed RBA is competitive or slightly outperforms BA \cite{BA_CVPR15}, especially on Oxford5k dataset. The proposed  SAH improves other methods by a fair margin. The improvement is more clear on Holidays and Oxford5k, e.g., SAH outperforms the most competitor RBA 2\%-3\% mAP at all code lengths. 

\subsection{Experiments with CNN feature maps}
Recently, in \cite{DBLP:journals/corr/ToliasSJ15,DBLP:conf/iccv/BabenkoL15,DBLP:journals/corr/AzizpourRSMC14} the authors showed that the activations from the convolutional layers of a convolutional neural network (CNN) can be interpreted as local features describing image regions. 
Motivated by those works, in this section we perform the experiments in which  activations of a convolutional layer from a pre-trained CNN are used as an alternative to SIFT features. It is worth noting that our work is the first one that evaluates hashing on the image representation aggregated from convolutional features. 
Specifically, we extract the activations of the $5^{th}$ convolutional layer (the last convolutional layer) of the pre-trained VGG network \cite{Simonyan14c}. Given an image, the activations form a 3D tensor of $W\times H \times C$, where $C=512$ which is number of feature maps and $W=H=37$ which is spatial size of the last convolutional layer. By using this setting, we can consider that each image is represented by $1,369$ local feature vectors with dimensionality $512$. In \cite{DBLP:conf/iccv/BabenkoL15}, the authors showed that the convolutional features are discriminative,  hence the embedding step is not needed for these features. 
Therefore, we directly use the convolutional features as the input for the proposed SAH. In order to make a fair comparison between SAH and other hashing methods, we aggregate the convolutional features with GMP \cite{DBLP:conf/cvpr/MurrayP14} and use the resulted vectors as the input for compared hashing methods. 

\begin{figure*}[!t]
\centering
\vspace{-0.2cm}
\subfigure[Holidays]{
       \includegraphics[scale=0.33]{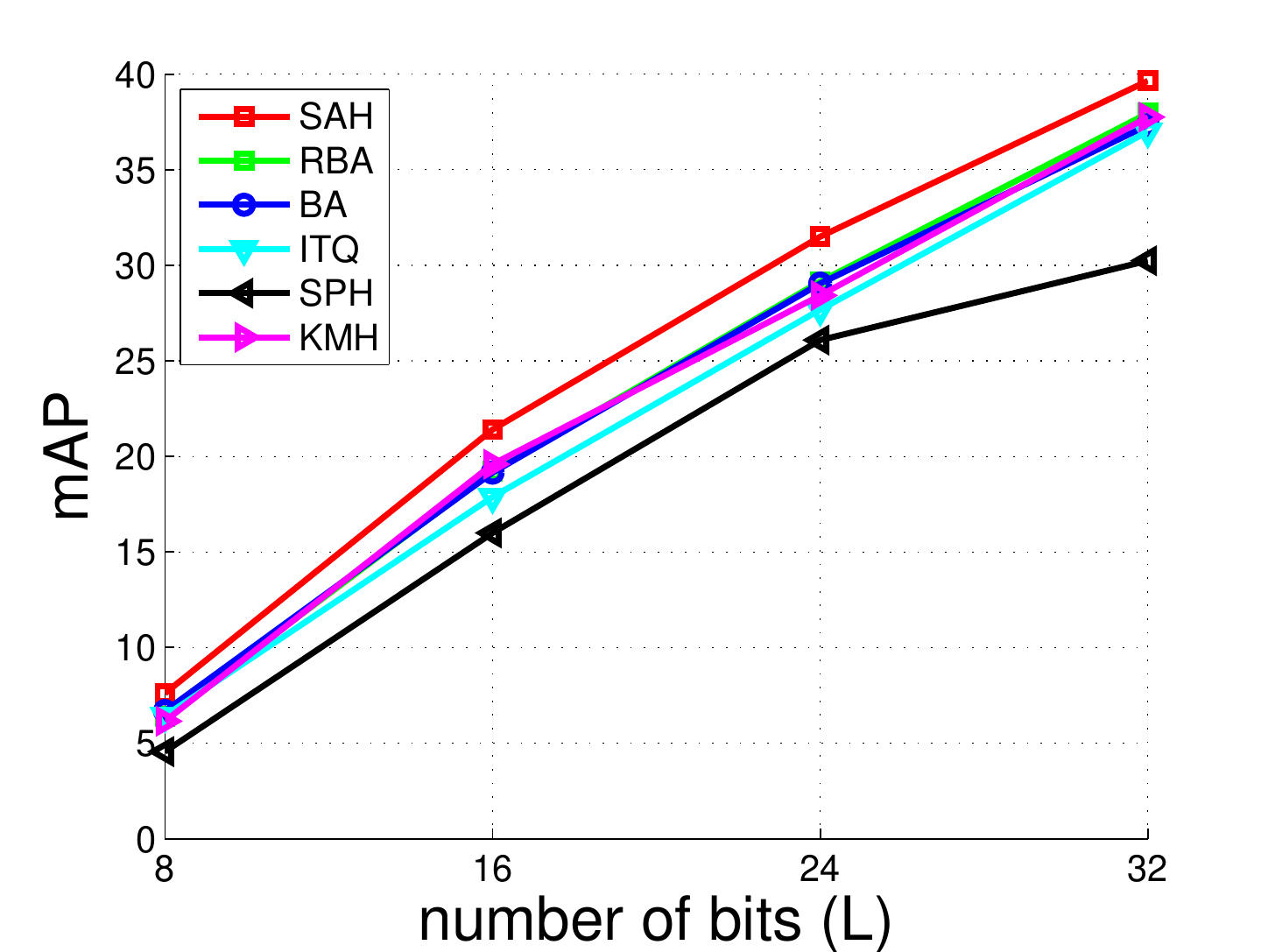}
       \label{fig:holidays_mAP_conv}
}
\subfigure[Oxford5k]{
       \includegraphics[scale=0.33]{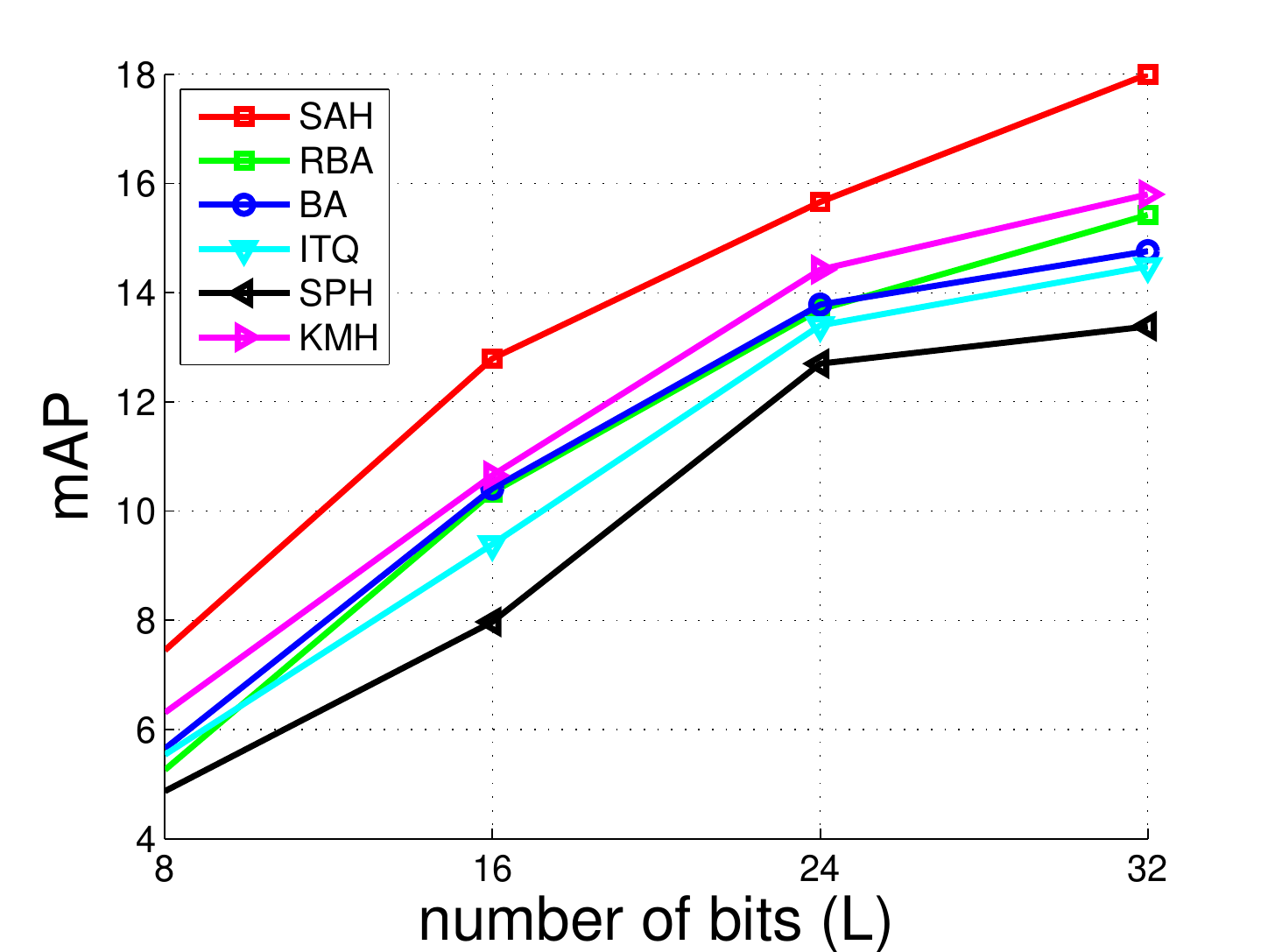} 
       \label{fig:oxford5k_mAP_conv}
}
\subfigure[Holidays+Flickr100k]{ 
       \includegraphics[scale=0.33]{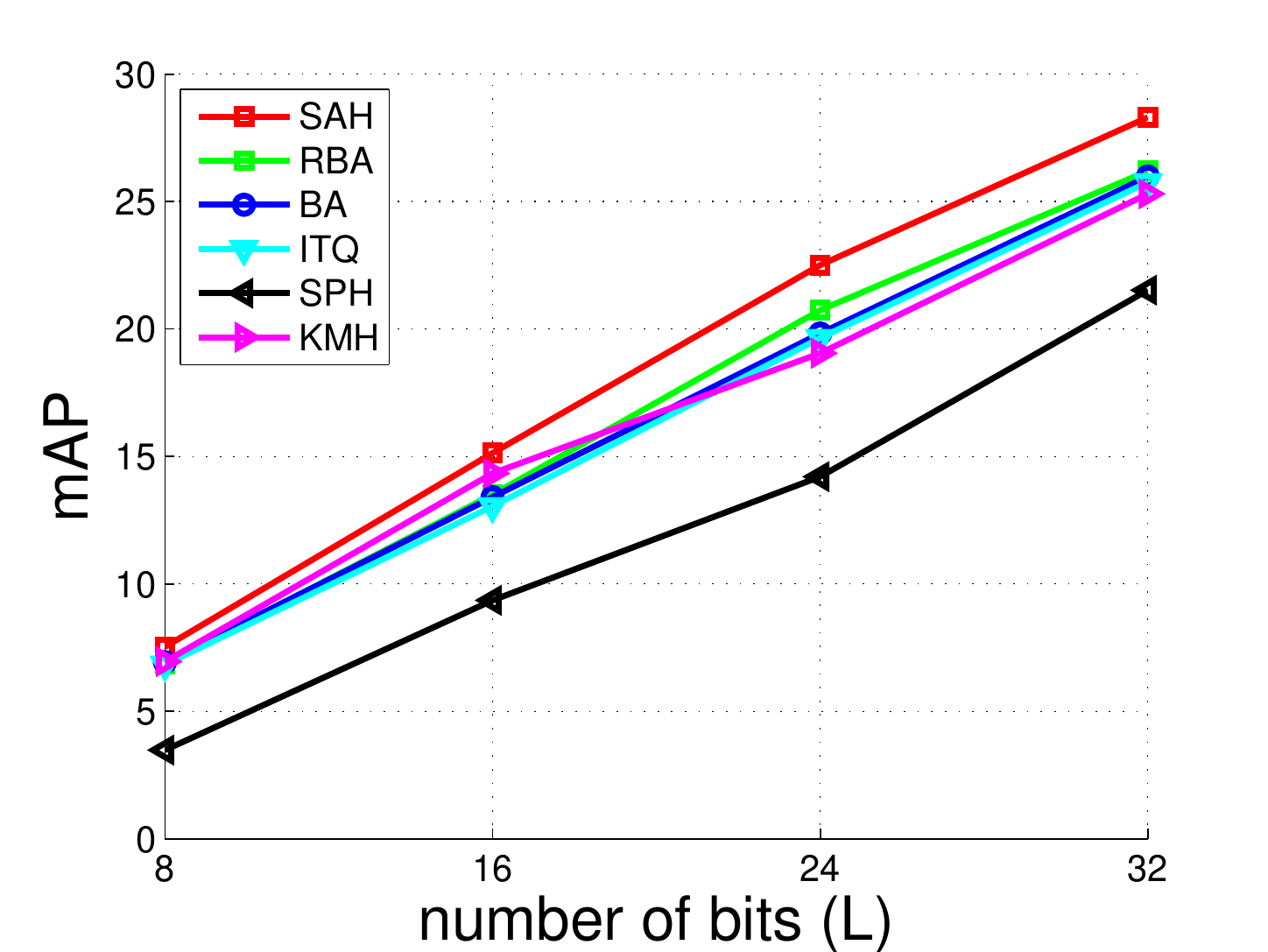} 
       \label{fig:holidays100k_mAP_conv}
}
\caption[]{mAP comparison between SAH and state-of-the-art unsupervised hashing methods when using convolutional features on Holidays, Oxford5k, and Holidays+Flickr100k }
\label{fig:lph-soa_CNN}
\end{figure*}

\vspace{-0.3cm}
\paragraph{Retrieval results}
Figure \ref{fig:lph-soa_CNN} shows the comparative mAP between methods. 
We can see BA \cite{BA_CVPR15}, KMH \cite{DBLP:conf/cvpr/HeWS13} and RBA achieve comparative results. It is clearly showed that the proposed SAH outperforms other methods by a fair margin. 
The improvements are more clear with longer code, e.g., SAH outperforms BA \cite{BA_CVPR15} 2\%-3\% mAP at $L = 32$ on three datasets. 
It is worth noting from Figure \ref{fig:lph-soa_CNN} and Figure \ref{fig:lph-soa} that at low code length, i.e., $L=8$, SIFT features and convolutional features give comparable results. However, when increasing the code length, the convolutional features significantly improves over the SIFT features, especially on Holidays and Holidays+Flickr100k datasets. For example, for SAH on Holidays+Flickr100k, the convolutional features improves mAP over the SIFT features about 5\%,  10\%, 14\% for $L=16,24$ and $32$, respectively.
\subsection{Comparison with fully-connected features}
In \cite{DBLP:conf/cvpr/RazavianASC14}, the authors showed that for image retrieval problem, using fully-connected features produced by a CNN outperforms most hand-crafted features such as VLAD \cite{herve_cvpr2010}, Fisher \cite{DBLP:conf/cvpr/PerronninD07}. 
In this section, we compare the proposed SAH 
with state-of-the-art unsupervised hashing methods which take the fully-connected features (e.g. outputs of the $7^{th}$ fully-connected layer from the pre-trained VGG network \cite{Simonyan14c}) as inputs. It is worth noting that there are few recent hashing methods which are based on end-to-end CNN, i.e., they jointly learn image representation and binary codes \cite{DBLP:conf/cvpr/LaiPLY15,DBLP:conf/cvpr/ZhaoHWT15,DBLP:journals/tip/ZhangLZZZ15}. However, those works are for supervised hashing and they are incomparable to this work which focuses on unsupervised hashing. For our proposed SAH, we take the convolutional features of the same pre-trained VGG network as inputs to demonstrate the 
benefit of the jointly learning of aggregating and hashing.


\begin{figure*}[!t]
\centering
\subfigure[Holidays]{
       \includegraphics[scale=0.33]{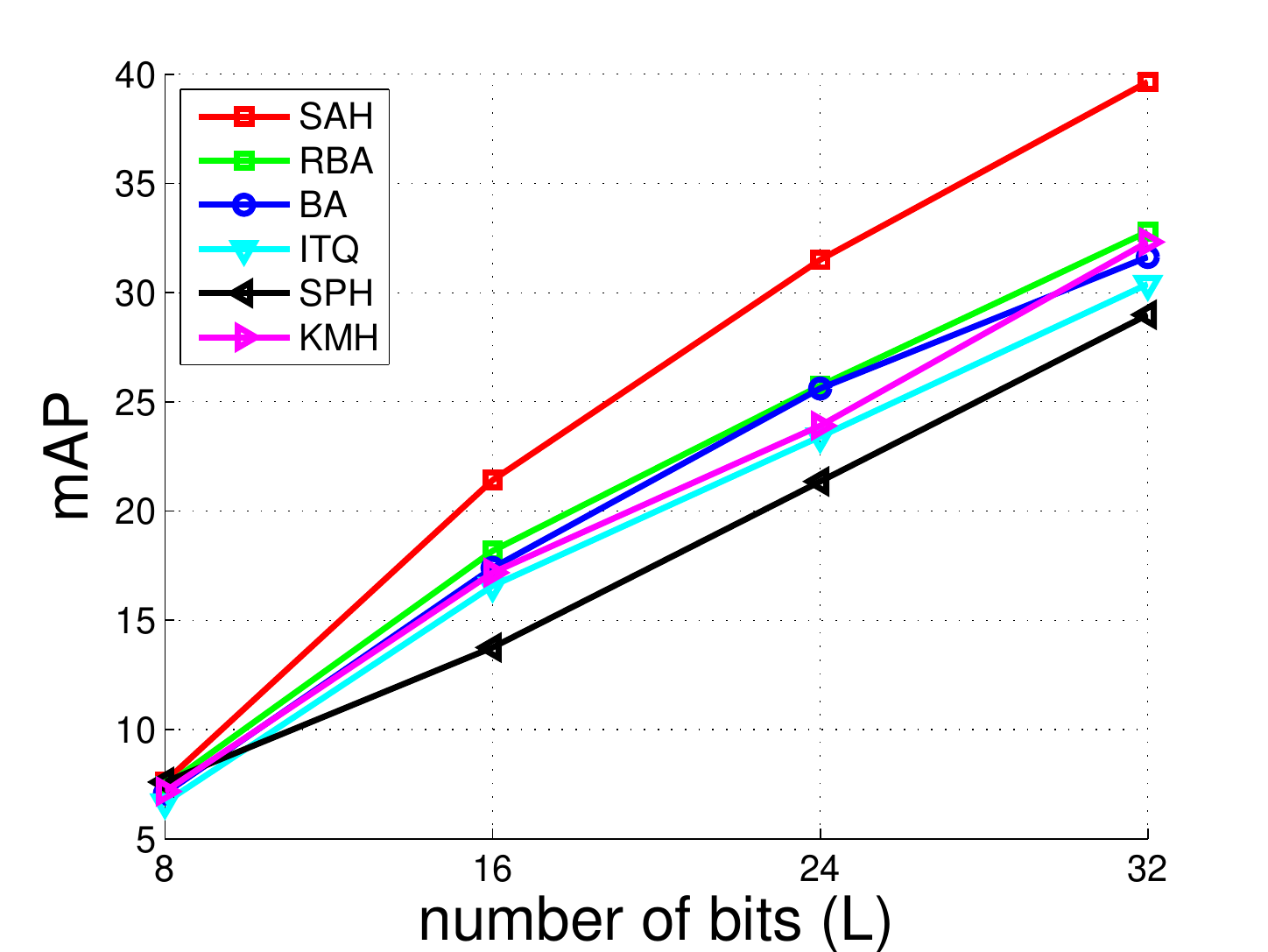}
       \label{fig:holidays_mAP_FC7}
}
\subfigure[Oxford5k]{
       \includegraphics[scale=0.33]{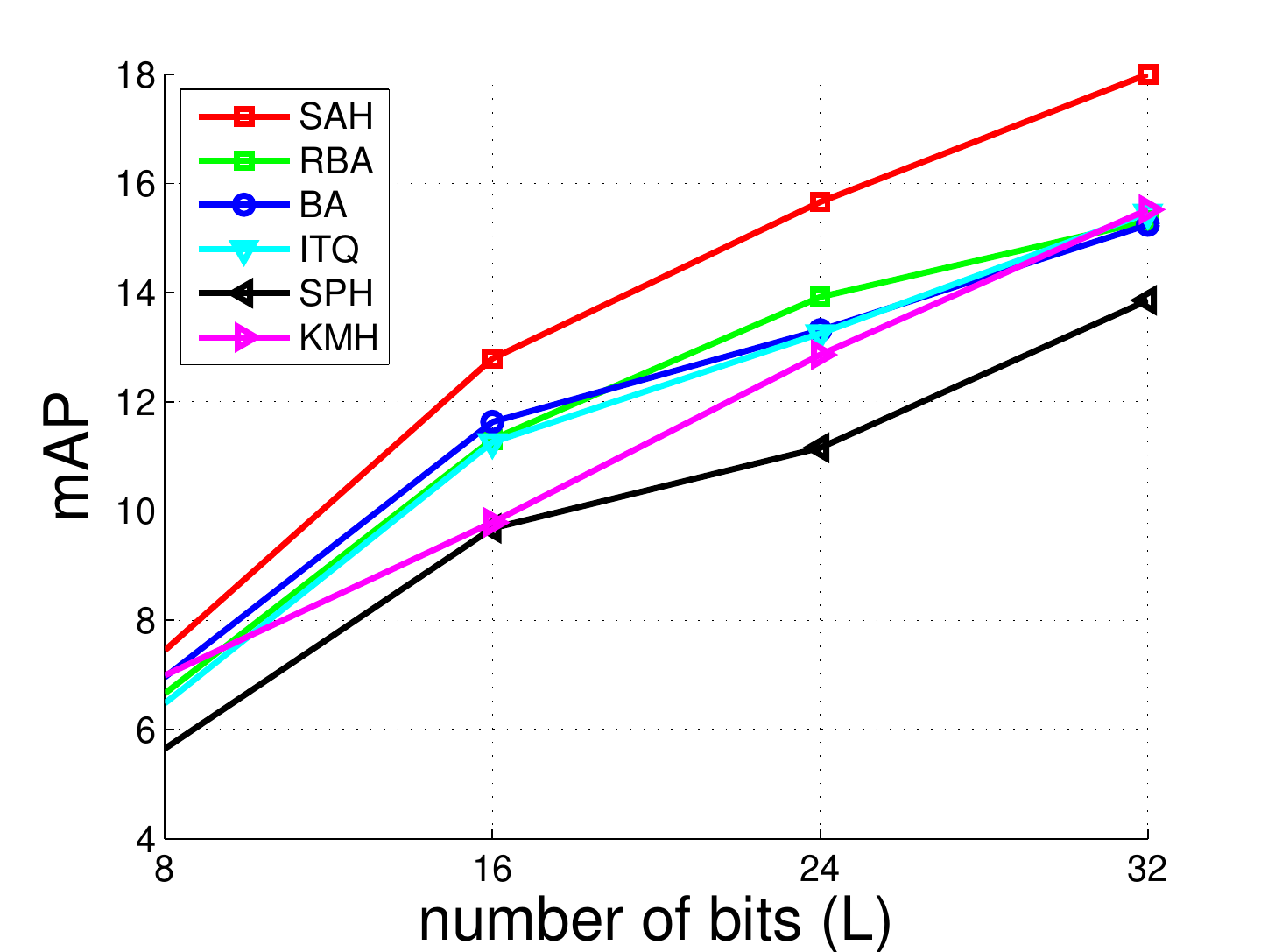} 
       \label{fig:oxford5k_mAP_FC7}
}
\subfigure[Holidays+Flickr100k]{ 
       \includegraphics[scale=0.33]{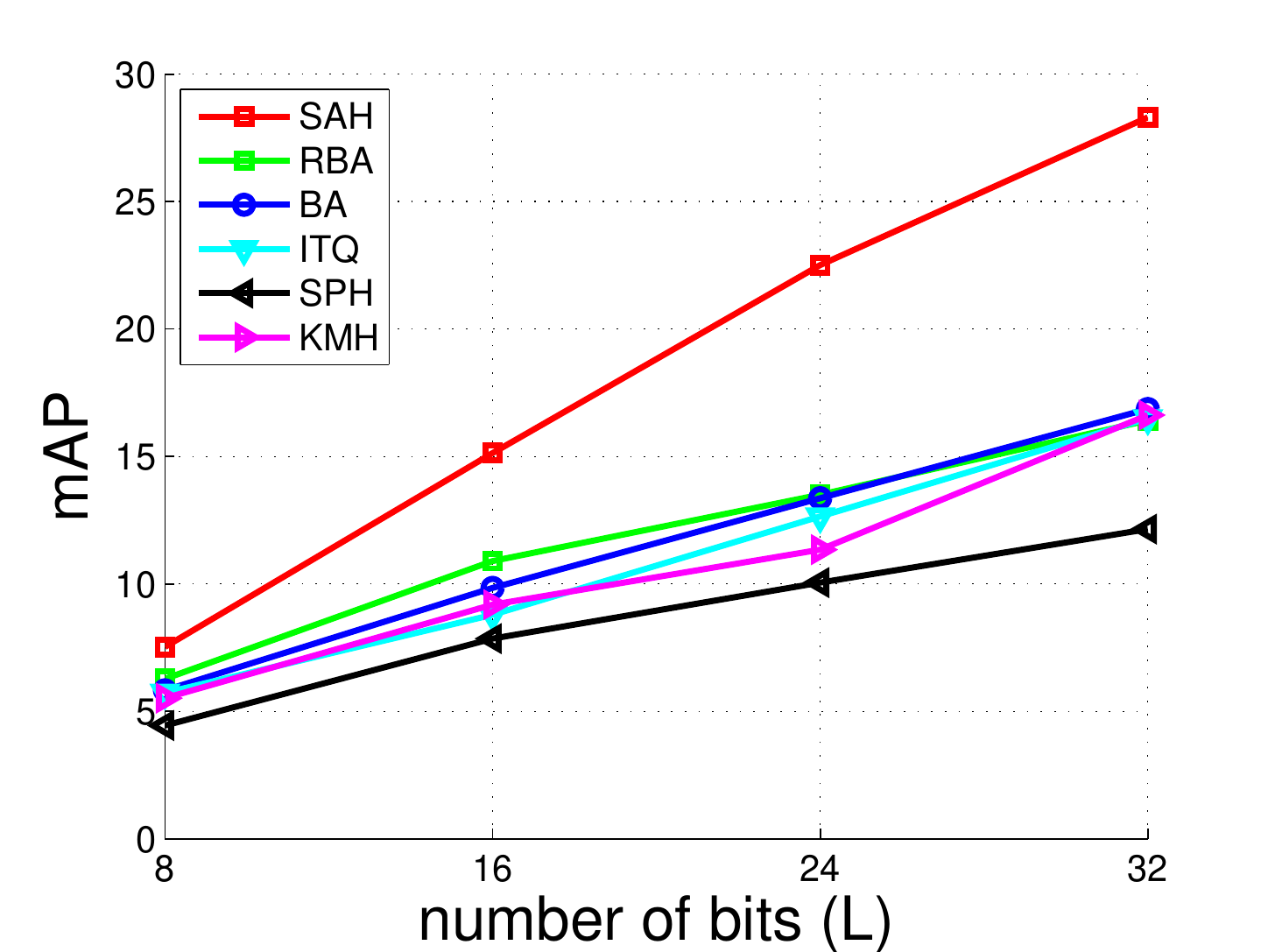} 
       \label{fig:holidays100k_mAP_FC7}
}
\caption[]{mAP comparison between SAH and state-of-the-art unsupervised hashing methods using fully-connected features on Holidays, Oxford5k, and Holidays+Flickr100k}
\label{fig:lph-soa_CNN_FC7}
\end{figure*}
\vspace{-0.3cm}
\paragraph{Retrieval results}
Figure \ref{fig:lph-soa_CNN_FC7} presents comparative mAP between methods. At low code length, i.e. $L=8$, SAH is competitive to other methods. However, when increasing the code length, SAH outperforms compared methods a large margin. The significant improvements are shown on Holidays and Holidays+Flickr100k datasets, e.g., at $L=32$, the improvements of SAH over BA \cite{BA_CVPR15} are 8\% and 11.4\% on Holidays and Holidays+Flickr100k, respectively. 
\vspace{-0.7cm}
\paragraph{Comparison with DeepBit \cite{deepbit2016}}
Recently, in \cite{deepbit2016}, the authors  proposed an end-to-end CNN-based unsupervised hashing approach. To the best of our knowledge, this is the only work using end-to-end CNN for unsupervised hashing. Starting with the pre-trained VGG network \cite{Simonyan14c}, they replaced the softmax layer of VGG with their binary layer 
and enforced several criteria on the binary codes learned at the binary layer, i.e., binary codes should: minimize the quantization loss with the output of the last VGG's fully connected layer, be distributed evenly, be invariant to rotation. 
Their network is fine-tuned using 50k training samples of CIFAR10. Note that as their approach is unsupervised, no label information is used during fine-tuning. Their comparative  mAP of the top $1,000$ returned images (with the class labels ground truth) on the testing set of CIFAR10 is cited in the top part of Table \ref{tab:deepbit}. 
\begin{table}[!t]
   \centering
   \footnotesize
   \caption{Comparison between DeepBit \cite{deepbit2016} and other unsupervised hashing methods on CIFAR10. The results in the first four rows are cited from \cite{deepbit2016}, which we have also reproduced.} 
    \begin{tabular}{|c|c|c|c|c|c|} 
    \hline
    Method    &16 bits &32 bits &64 bits\\ \hline
    ITQ \cite{DBLP:conf/cvpr/GongL11} &15.67 &16.20  &16.64\\ \hline
    KMH \cite{DBLP:conf/cvpr/HeWS13} &13.59 &13.93  &14.46\\ \hline
    SPH \cite{CVPR12:SphericalHashing} &13.98 &14.58  &15.38\\ \hline
    DeepBit \cite{deepbit2016} &19.43 &24.86 &27.73\\ \hline \Xhline{3\arrayrulewidth}
    ITQ-CNN &38.52	&41.39 &44.17\\ \hline
    KMH-CNN &36.02	&38.18 &40.11\\ \hline
    SPH-CNN &30.19	&35.63 &39.23\\ \hline
    SAH &\textbf{41.75} &\textbf{45.56} &\textbf{47.36}\\ \hline
    \end{tabular}
    \label{tab:deepbit}
\end{table}
Note that their reported results of ITQ, KMH, SPH 
come from \cite{Liong_2015_CVPR} in which GIST features are used. 
Therefore, we also evaluate those three hashing methods on the features extracted from the activations of the last fully connected layer of the same pre-trained VGG \cite{Simonyan14c} (without fine-tuning). These results, i.e. ITQ-CNN, KMH-CNN, SPH-CNN, are presented in the bottom part of Table \ref{tab:deepbit}. It clearly shows that 
ITQ-CNN, KMH-CNN, SPH-CNN have significant improvements (using fully-connected instead of GIST). In order to evaluate the proposed SAH, we extract the activations of the last convolutional layer of the same pre-trained VGG and use them as input. 
The results of SAH presented in the last row in Table \ref{tab:deepbit} show that at the same code length, SAH significantly outperforms the recent end-to-end work DeepBit \cite{deepbit2016}, i.e., the mAP improvements are 22.3\%, 20.7\%, 19.6\% at $L=16$, $32$ and $64$, respectively. Furthermore, SAH also outperforms ITQ-CNN, KMH-CNN, SPH-CNN with a fair margin. 

\section{Conclusion}
\label{sec:concl}
In this paper, we first introduce Relaxed Binary Autoencoder (RBA) hashing method in which instead of solving the hard binary constraint, we minimize the binary quantization loss. 
Compare to Binary Autoencoder, the proposed RBA achieves not only faster training but also competitive retrieval results.
We then propose a novel unsupervised hashing approach called SAH by integrating feature aggregating and hash function learning into a joint optimization framework. Extensive experiments on benchmark datasets with SIFT, convolutional, and fully-connected features demonstrate that the proposed SAH method outperforms state-of-the-art unsupervised hashing methods.

{\small
\bibliographystyle{ieee}
\bibliography{hash}
}

\end{document}